\documentclass[acmtog]{acmart}
\usepackage{adjustbox}
\usepackage{array}
\usepackage{gensymb}
\usepackage{multirow}

\usepackage[normalem]{ulem}
\usepackage{graphics}
\usepackage{array}
\usepackage{multirow}
\usepackage{wrapfig}
\usepackage{hhline}
\usepackage{pifont}%

\DeclareMathOperator*{\argmin}{arg\,min} %
\newcolumntype{K}[1]{>{\centering\arraybackslash}p{#1}}

\usepackage{duckuments}
\usepackage{tikzducks}
\usepackage{graphicx}
\usepackage{adjustbox}
\usepackage{pifont}

\usepackage{float}

\usepackage{atbegshi}
\newcommand{\placetextbox}[3]{
  \setbox0=\hbox{#3}%
  \AtBeginShipoutNext{\AtBeginShipoutUpperLeft{%
    \put(\dimexpr#1\paperwidth\relax,-\dimexpr#2\paperheight\relax)
    {\vtop{{\null}\makebox[0pt][c]{#3}}}%
  }}%
}

\usepackage{lipsum}

\usepackage[normalem]{ulem}
\useunder{\uline}{\ul}{}

\newcommand{\ua}{\uparrow}
\newcommand{\da}{\downarrow}

\AtBeginDocument{%
  \providecommand\BibTeX{{%
    \normalfont B\kern-0.5em{\scshape i\kern-0.25em b}\kern-0.8em\TeX}}}

\setcopyright{acmlicensed}
\acmJournal{TOG}
\acmYear{2023} \acmVolume{1} \acmNumber{1} \acmArticle{1} \acmMonth{1} \acmPrice{15.00}\acmDOI{10.1145/3630750}

\acmJournal{TOG}

\begin{document}

\title{Intrinsic Image Decomposition via Ordinal Shading}

\author{Chris Careaga}
\author{Ya\u{g}{\i}z Aksoy}
\affiliation{
  \institution{Simon Fraser University}
  \city{Burnaby}
  \state{BC}
  \country{Canada}
}

\renewcommand{\shortauthors}{Careaga and Aksoy}

\begin{abstract}

Intrinsic decomposition is a fundamental mid-level vision problem that plays a crucial role in various inverse rendering and computational photography pipelines. Generating highly accurate intrinsic decompositions is an inherently under-constrained task that requires precisely estimating continuous-valued shading and albedo. In this work, we achieve high-resolution intrinsic decomposition by breaking the problem into two parts. First, we present a dense ordinal shading formulation using a shift- and scale-invariant loss in order to estimate ordinal shading cues without restricting the predictions to obey the intrinsic model. We then combine low- and high-resolution ordinal estimations using a second network to generate a shading estimate with both global coherency and local details. We encourage the model to learn an accurate decomposition by computing losses on the estimated shading as well as the albedo implied by the intrinsic model. We develop a straightforward method for generating dense pseudo ground truth using our model’s predictions and multi-illumination data, enabling generalization to in-the-wild imagery. We present exhaustive qualitative and quantitative analysis of our predicted intrinsic components against state-of-the-art methods. Finally, we demonstrate the real-world applicability of our estimations by performing otherwise difficult editing tasks such as recoloring and relighting. 
\end{abstract}

\ccsdesc[500]{Computing methodologies~Image representations}
\ccsdesc[500]{Computing methodologies~Image manipulation}

\keywords{intrinsic decomposition, inverse rendering, mid-level vision, shading and reflectance estimation, image manipulation}

\newcommand{\chris}[1]{\textcolor{red}{{[chris: #1]}}}
\newcommand{\yagiz}[1]{\textcolor{green}{{[Ya\u{g}{\i}z: #1]}}}
\newcommand{\revision}[1]{{#1}}

\begin{teaserfigure}
  \includegraphics[width=\linewidth]{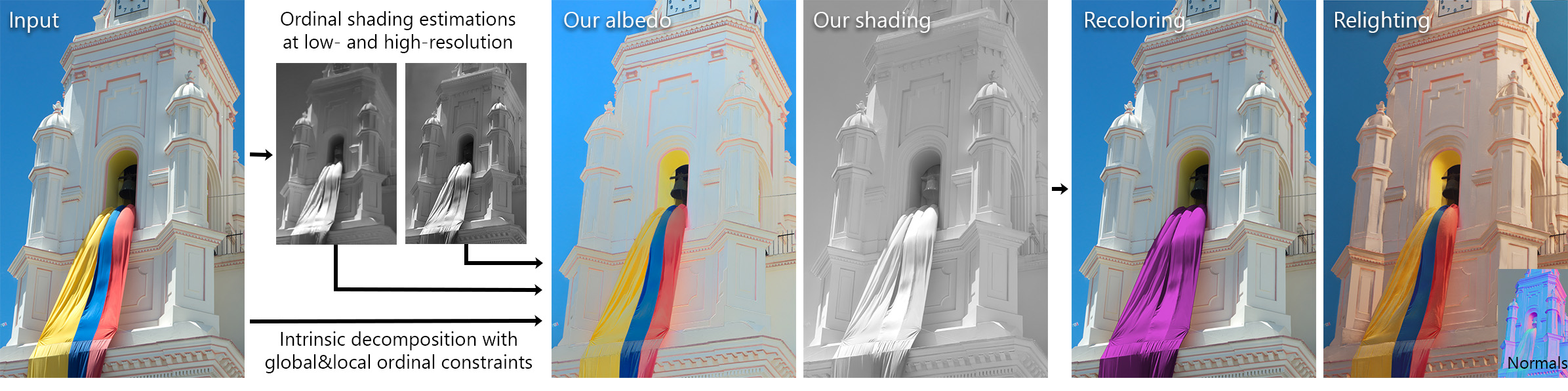}
  \\
  \includegraphics[width=\linewidth]{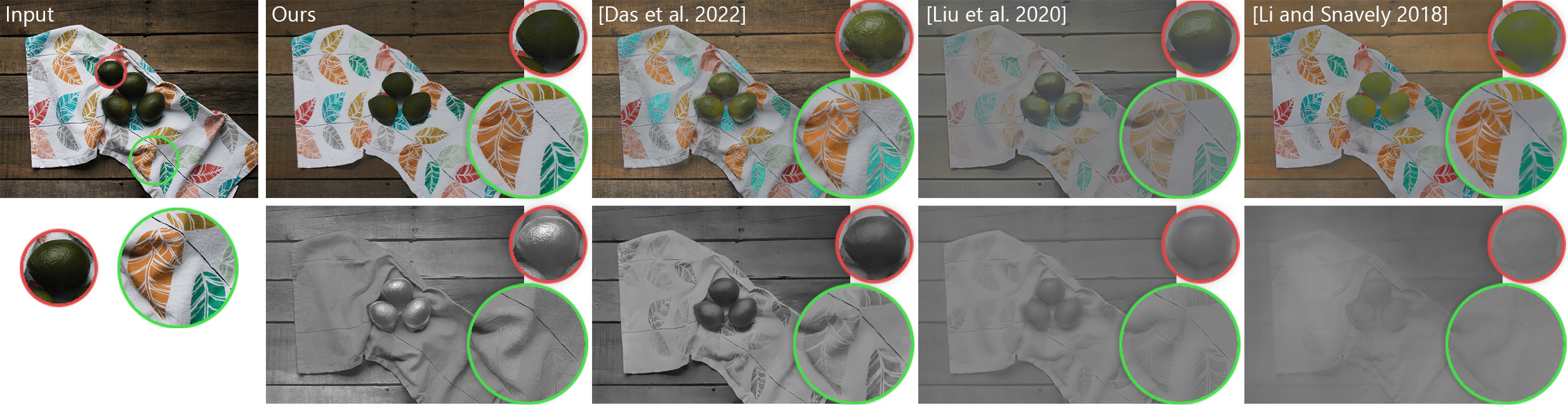}
  \caption{(Top) We propose a two-step pipeline for intrinsic decomposition. We first estimate low- and high-resolution ordinal shading maps that provide global and local constraints. We then estimate the full intrinsic decomposition using these ordinal inputs. Our decomposition results can be used for applications like recoloring and relighting. (Bottom) When compared to prior works, our method generates high-quality results on challenging images in the wild without leaking textures between each component and accurate shading values around specularities. \hfill \footnotesize{Images from Unsplash by Miguel Ibáñez (top) and Debby Hudson.}}
  \label{fig:teaser}
\end{teaserfigure}

\maketitle

\placetextbox{0.14}{0.03}{\includegraphics[width=4cm]{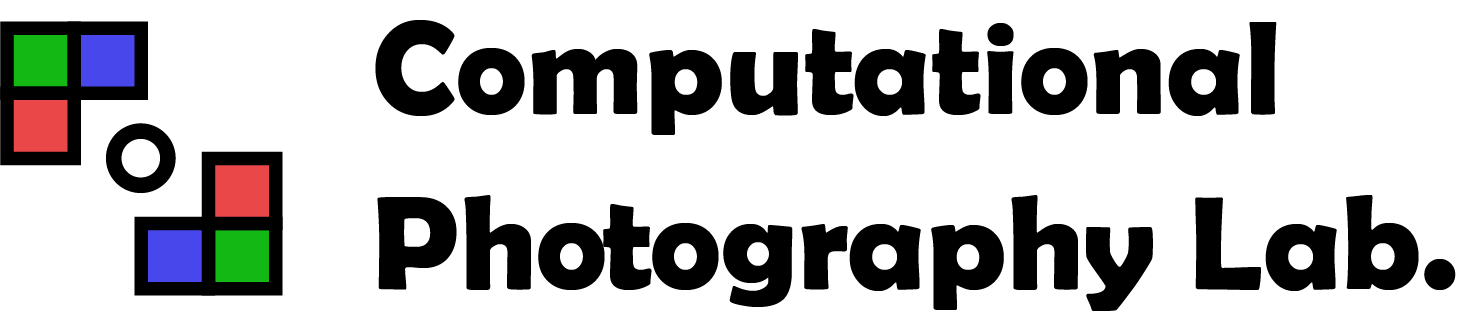}}
\placetextbox{0.85}{0.03}{Find the project web page here:}
\placetextbox{0.85}{0.045}{\textcolor{purple}{\url{https://yaksoy.github.io/intrinsic/}}}

\section{Introduction}
\label{sec:intro}

\begin{figure*}
    \centering
    \includegraphics[width=\linewidth]{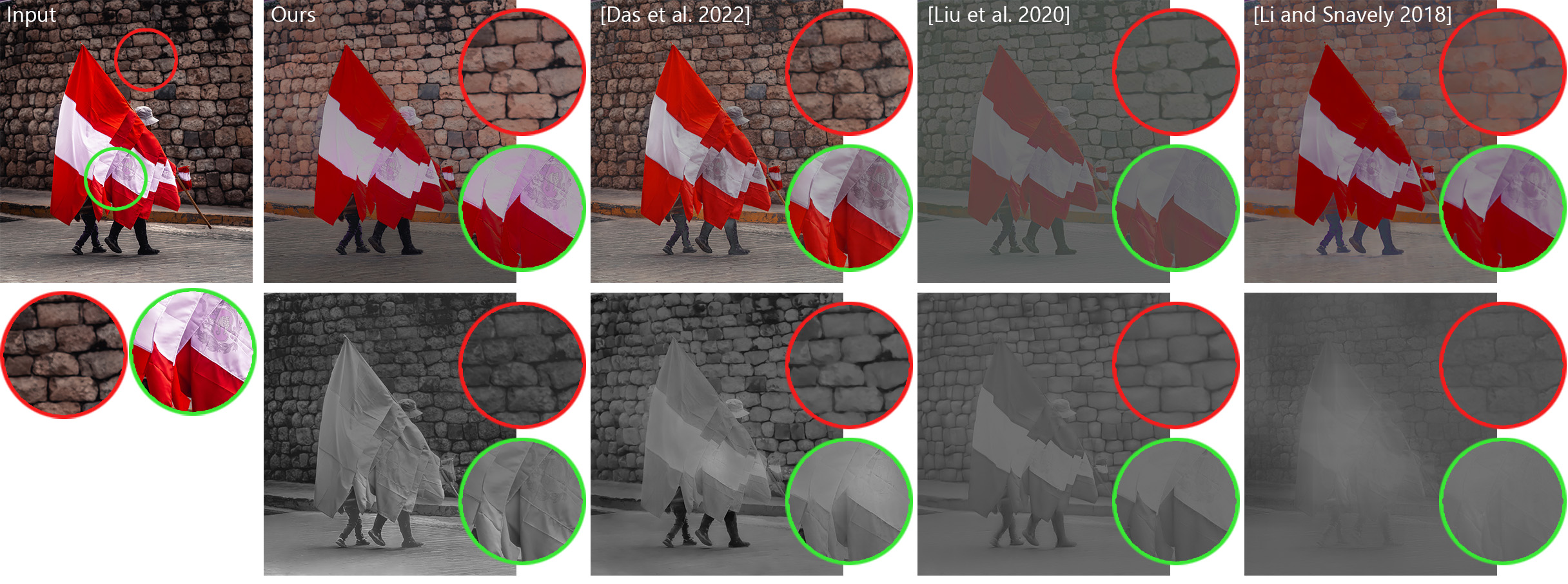}
    \\
    \includegraphics[width=\linewidth]{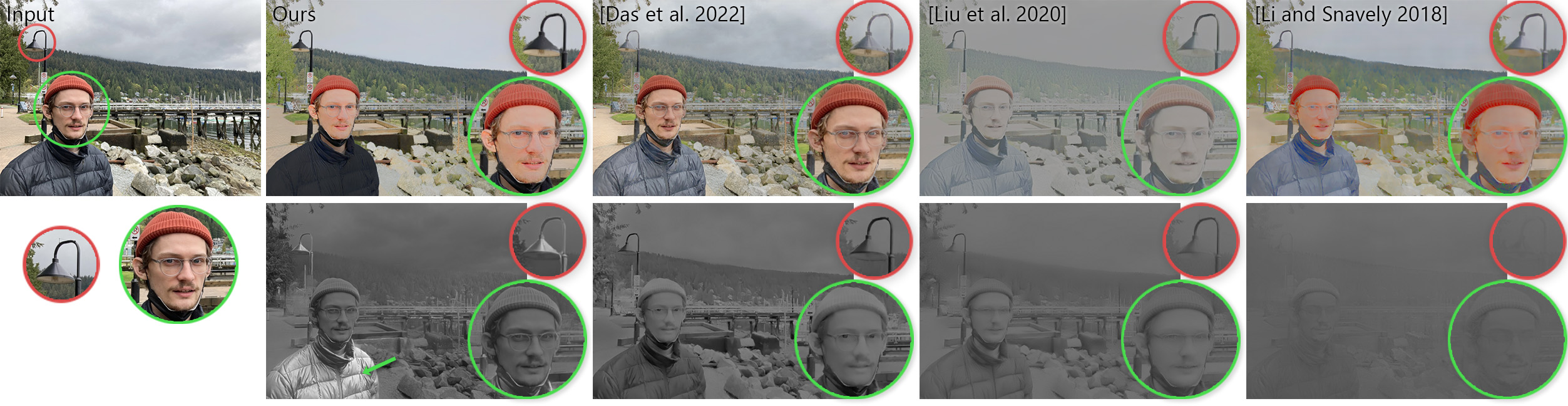}
    \caption{
    Our method is able to generate accurate shading and albedo for high-resolution in-the-wild imagery. When compared to prior approaches, our method is able to properly predict intrinsic components for images outside of the training distribution such as human faces. Our estimations are also accurate in difficult regions where both albedo and shading vary without leaking textures between each component. Thanks to our formulation, we are the only method that can accurately predict shading values on specular surfaces such as the black jacket and the light post. \hfill \footnotesize{Top image from Unsplash by Mauro Lima.}
    }
    \label{fig:introComparison}
\end{figure*}

Intrinsic image decomposition is a fundamental mid-level vision problem that aims to represent an image as the product of the reflectance of the materials and the effect of illumination in the scene:
\begin{equation}
    I = A * S,
    \label{eq:intrinsicmodel}
\end{equation}
where $I$, $A$, and $S$ represent the input image, the (Lambertian) albedo, and the shading, respectively.
Since intrinsic decomposition separates the illumination-invariant scene properties from the illumi- nation-dependent lighting effects, it is a critical component for a wide range of computational photography pipelines such as relighting, recoloring, and compositing. 
Realistic image editing through intrinsics requires accurate decompositions at high resolutions for in-the-wild photographs. 
Prior data-driven methods have not been able to live up to these requirements \cite{garces2022survey, bonneel2017intrinsic} and as a result, intrinsic computational photography methods have not yet been widely adopted by digital artists.

The intrinsic model in Equation~\ref{eq:intrinsicmodel} is inherently under-constrained, as well as scale-invariant -- i.e. for a given A and S, $\frac{1}{c} A$ and $cS$ also satisfy the model for all $c > 0$.
Shading is a continuous-valued map that represents the complex interactions between the light sources, the 3D geometry, and the material properties present in the scene. 
This makes intrinsic decomposition a high-level problem wherein neural networks have to rely on contextual information in the scene. 
These challenges are coupled with the lack of dense ground-truth data on real-world images. 
Due to the complexity of the problem, state-of-the-art intrinsic decomposition models fail to produce accurate results at high resolutions in the wild.
\begin{figure*}
    \centering
    \includegraphics[width=\linewidth]{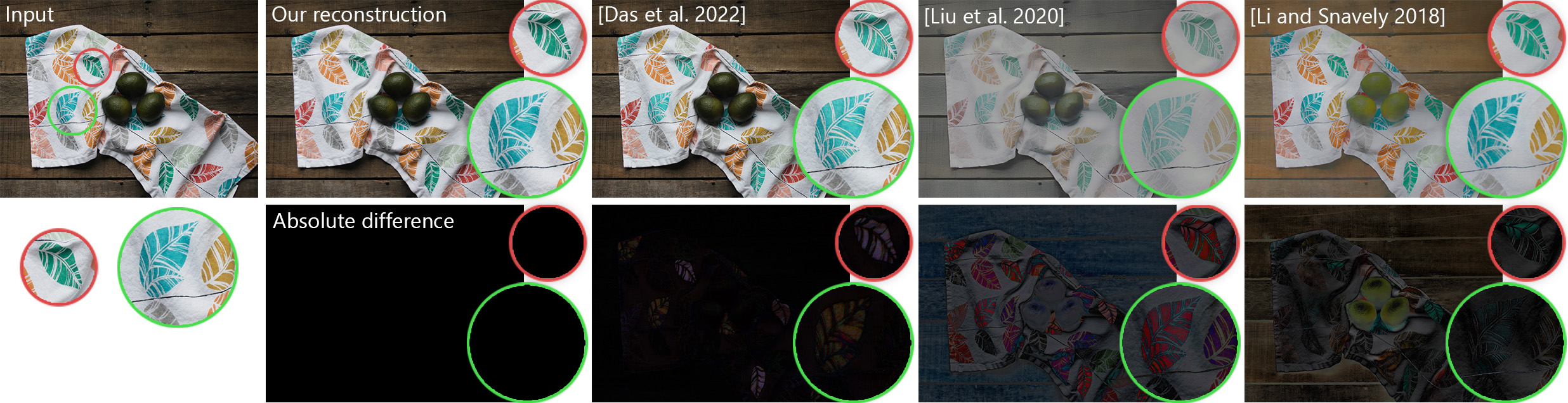}
    \caption{
    Most deep learning approaches separately predict albedo and shading components, only encouraging reconstruction via losses between the input image and the combined intrinsic components. 
    In contrast, our method only predicts shading and uses the intrinsic image formulation to yield the implied albedo component. 
    This formulation ensures perfect reconstruction which is necessary for image editing applications.  \hfill \footnotesize{Image from Unsplash by Debby Hudson.}
    }
    \label{fig:introComparisonReconst}
\end{figure*}
In this paper, we achieve high-resolution intrinsic decomposition by breaking the problem into two. 
We first develop a scale- and shift-invariant (SSI) dense shading estimation formulation that we call \emph{ordinal shading}. 
In this formulation, we relax the constraint to satisfy the intrinsic model while enforcing the estimation of reliable ordinal relationships between pixels in the dense output. 
We show that our simplified ordinal problem definition makes it possible to estimate shading smoothness and discontinuities at high resolutions. 

In the second step, we make use of two dense ordinal shading maps generated at two different resolutions. 
The low-resolution ordinal shading is generated at the receptive field resolution of the ordinal network, providing global ordinal constraints. 
The second ordinal shading is generated at a much higher resolution to provide highly detailed shading discontinuities as local constraints. 
We feed these two ordinal maps together with the image to a second network that enforces the intrinsic model through losses on both albedo and shading. 
With high resolution ordinal constraints readily available to the network, we show that we can generate highly detailed intrinsic decompositions that can be used in image editing tasks. 

Intrinsic decomposition networks are typically trained on synthetic data with some real-world data with sparse ground-truth annotations.
To train our networks with high-resolution real-world dense ground truth data, we derive a dataset from the Multiple Illuminations Dataset by Murmann et al.~\shortcite{murmann19multi}. 
By exploiting the fact that albedo is illumination-invariant, and hence a constant across changing illumination, we formulate a robust pseudo-ground-truth generation method using the 25 images per scene with different lighting conditions provided in this dataset. 
We show that we can generalize our high-resolution intrinsic decomposition method to in-the-wild examples using dense training on real photographs. 

Our method can generate highly detailed intrinsic decomposition results in the wild. 
We can generate smooth shading results on heavily textured surfaces such as the cloth in Figure~\ref{fig:teaser} and on the flag in Figure~\ref{fig:introComparison} while reconstructing the input image faithfully as shown in Figure~\ref{fig:introComparisonReconst}. 
Our formulation is robust against challenging regions in the image such as specularities as seen on the avocadoes in Figure~\ref{fig:teaser} and the lamp post in Figure~\ref{fig:introComparison}. 
We can also generalize to cases not represented in our training set such as human faces as Figure~\ref{fig:introComparison} shows. 
We present extensive evaluations of our method against the state-of-the-art qualitatively and quantitatively and show that we improve the performance in terms of albedo sparsity, shading smoothness, and sharpness of shading discontinuities in a variety of scenarios. 
Our high-resolution intrinsic decomposition method enables realistic image editing applications in the wild such as recoloring and relighting as shown in Figure~\ref{fig:teaser}. 

\begin{figure*}
    \centering
    \includegraphics[width=\linewidth]{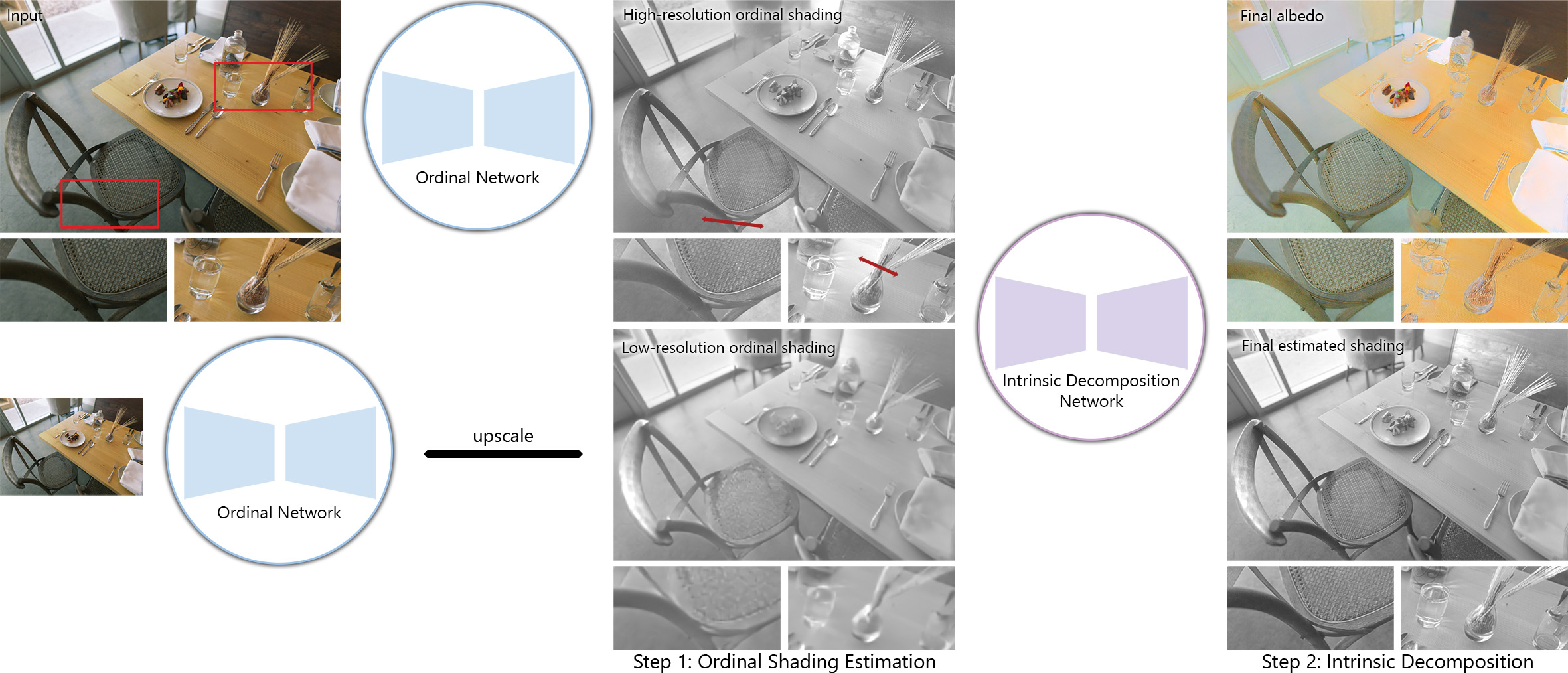}
    \caption{
    We achieve intrinsic decomposition in two steps. 
    In the first step, we generate two ordinal shading estimations, one at the receptive field resolution of our network, and another at a much higher resolution. 
    The low-resolution estimation provides globally coherent ordinal constraints but it lacks high-resolution details. 
    The high-resolution estimation, on the other hand, contains highly detailed shading discontinuities providing us with reliable local constraints. 
    However, it may have inconsistencies across distant image regions as visible on the two sides of the glass in the bottom inset. 
    We utilize these two ordinal estimations as input to our second network together with the original input image. 
    With global and local constraints readily provided to the network, we are able to generate a globally consistent shading with high-resolution details and sharp shading discontinuities. 
    We then compute the corresponding albedo using the input image and the estimated shading using the intrinsic equation. 
      \hfill \footnotesize{Image from Unsplash by Erik Binggeser.}
    }
    \label{fig:pipeline}
\end{figure*}

\section{Related Work}
\label{sec:related}

Due to its wide range of application scenarios, intrinsic image decomposition has generated a lot of attention in the computer vision and computational photography literature. While earlier methods focus on developing low-level priors for shading and albedo, with increased availability of training data, the focus of the field first shifted to sparse ordinal representations and then to direct regression of the continuous-valued shading. In this section, we discuss the related ordinal and data-driven techniques in the literature and refer the reader to the recent survey on intrinsic decomposition by Garces et al.~\shortcite{garces2022survey} for a comprehensive overview.

\paragraph{Ordinality in intrinsic decomposition}
The first relatively large-scale ground-truth datasets came in the form of sparse ordinal annotations on real-world images. The Intrinsic Images in-the-Wild (IIW)~\cite{bell2014intrinsic} dataset provides relative annotations of albedo brightness between sparsely sampled pixel pairs. Following the release of IIW, multiple works focused on the simpler \emph{ordinal} definition of the problem where data-driven systems are trained to predict ordinal relationships between the albedos of pixel pairs~\cite{zoran2015learning, zhou2015learning, narihira2015learning}.  The estimated sparse ordinal relationships can then be used to estimate the dense decomposition via MRF-based optimization~\cite{zhou2015learning} or by solving a linear system ~\cite{zoran2015learning}.

We also focus on the ordinal definition of the problem in this paper. Rather than predicting sparse ordinal relationships on albedo, however, we define a novel dense ordinal shading space that we train with dense ordinal losses. Our method has some parallels to the works of Zoran et al.~\shortcite{zoran2015learning} and Zhou et al.~\shortcite{zhou2015learning}.  Both of these methods estimate ordinal relationships using both global and local image information. They then use these estimated ordinal albedo constraints to regress the dense decomposition result using traditional optimization techniques. Rather than directly estimating ordinal relationships using global and local inputs, we propose to generate two separate estimations that capture this information. We generate global constraints by resizing the image to fit the receptive field of our network, and local constraints by estimating ordinal shading at a much higher resolution. We then use these ordinal estimations as input to a second network to generate a high-resolution decomposition.

\paragraph{Data-driven approaches}
As physically-based rendering techniques have improved, it has become feasible to train intrinsic decomposition networks using rendered datasets. Many data-driven approaches propose to train CNN models using direct supervision on small-scale datasets \cite{shi2016learning, janner2017intrinsic, meka2018lime, baslamisli2018cnn, ma2018single} like ShapeNet \revision{\cite{shapenet2015}}, MPI Sintel \revision{\cite{MPISintel}} or the MIT Dataset \revision{\cite{grosse2009ground}}. The CGIntrinsics Dataset~\cite{li2018cgintrinsics} is the first large-scale dataset of rendered scenes with ground-truth intrinsics. With its introduction, many approaches have been proposed that either utilize the CGIntrinsics ground-truth or generate their own similar rendered datasets~\cite{luo2020niid, zhou2019glosh, li2020inverse, liu2020unsupervised, sengupta19neural, zhu2022irisformer}. These works depend on scale-invariant losses to directly regress the shading and albedo. Due to the complexity of the intrinsic decomposition problem and the limited capacity of neural networks, these methods fail to generate accurate and high-resolution estimations. By first making use of the simpler ordinal definition of the problem followed by regression of full decomposition with ordinal constraints as input, we are able to generate high-resolution decompositions with sharp shading discontinuities and globally coherent sparse albedo maps.

Most prior data-driven models utilize architectures that estimate shading and albedo separately~\cite{narihira2015direct, shi2016learning, baslamisli2018cnn, cheng2018intrinsic, li2018cgintrinsics, li2018learning, zhou2019glosh, luo2020niid, das2022pie}. These methods enforce constraints on each intrinsic component and incorporate a reconstruction loss that favors outputs that reproduce the input image when multiplied. There is no guarantee that these methods will generate a faithful reconstruction of the input image for a novel scene, as shown in Figure~\ref{fig:introComparisonReconst}, which limits their use in image editing applications. We opt for an approach similar to that of \citet{fan2018revisiting} and \citet{lettry2018unsupervised,lettry2018darn} and derive the albedo using estimated shading and Equation \ref{eq:intrinsicmodel}. Since this process is differentiable, we jointly optimize for both albedo and shading using dedicated losses to each while using a single network and guaranteeing perfect image reconstruction.

\begin{figure*}
    \centering
    \includegraphics[width=\linewidth]{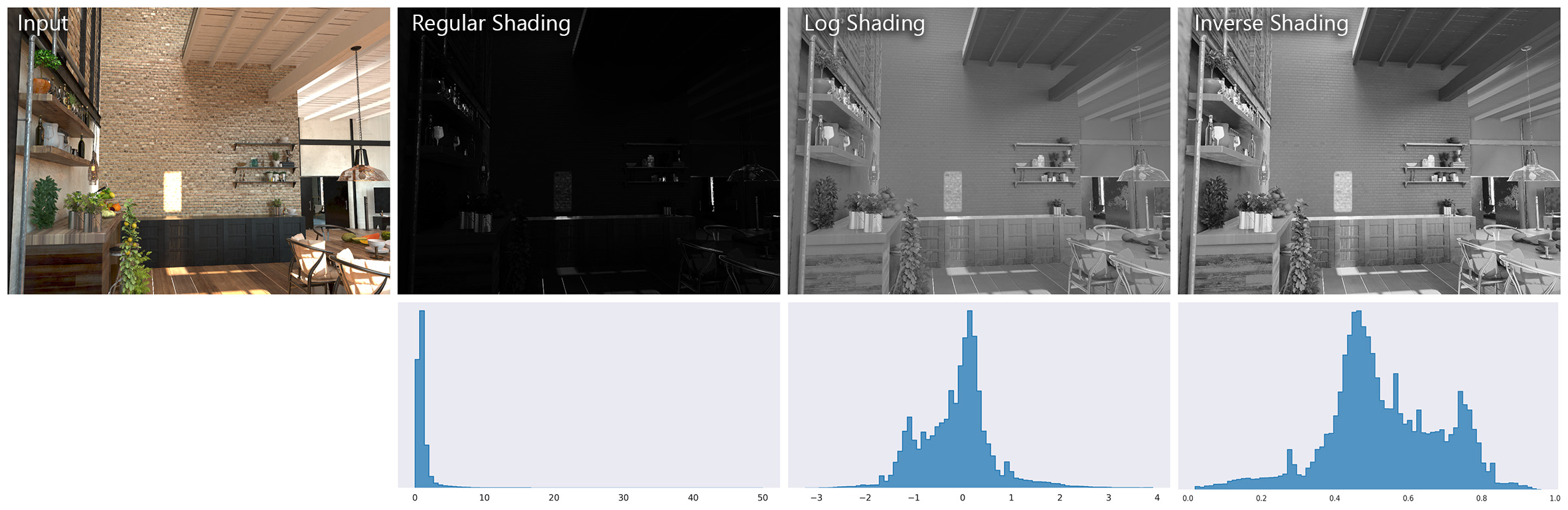}
    \caption{We visualize various shading representations for an image from the Hypersim dataset~\cite{roberts2021hypersim}. 
    The unaltered linear shading is dominated by specular outliers causing a long-tailed distribution.
    While the log shading has a more balanced distribution, it still lacks contrast in the mid-range values. 
    It also has an undefined range of possible shading values and a long tail due to the specularities in the scene. 
    Our proposed representation, inverse shading, best utilizes the available range of values and is guaranteed to be in $[0, 1]$. 
    The original and log-space representations are min-max normalized for visualization.}
    \label{fig:ordinal:shd_rep}
\end{figure*}

\paragraph{Real-world training datasets}
Although a number of synthetic datasets have been developed with ground-truth intrinsic components \cite{roberts2021hypersim, li2021openrooms, li2018cgintrinsics, krahenbuhl2018free, le2021eden}, it is still a difficult task to train models that generalize to real-world imagery. Rendered datasets typically depict homogeneous indoor scenes, with the exception of the GTA Dataset \cite{krahenbuhl2018free}, which contains outdoor scenes rendered using a video game engine. This leaves a \revision{domain} gap between the training data and in-the-wild photographs \cite{garces2022survey}. To address this shortcoming, many methods~\cite{li2018cgintrinsics, zhou2019glosh, fan2018revisiting} supplement their training procedure with the sparse annotations from the IIW~\cite{bell2014intrinsic} and SAW~\cite{kovacs17shading} datasets. 
IIW, together with SAW, which provides sparse annotations for shading smoothness, are currently the only real-world datasets with ground-truth information. 
However, as sparse data is only able to provide a weak supervision~\cite{garces2022survey}, their usefulness in generalizing to in-the-wild photographs is limited.
Other works leverage image sequences of stationary scenes under varying illumination~\cite{lettry2018unsupervised, li2018learning, ma2018single, bi2018deep}. These approaches all utilize similar loss functions that encourage reflectance consistency across multiple illuminations, but require complicated priors and do not guide the network toward a single ground-truth. In this work, we propose a method to generate pseudo-ground-truth intrinsic components from multi-illumination data using the Multiple Illuminations Dataset~\cite{murmann19multi}. This enables us to use a unified dense loss formulation on both synthetic and real data, which is a key step to bridging the intrinsic decomposition generalization gap.

\section{Method Overview and Preliminaries}
\label{sec:method}

In this work, we achieve high-resolution intrinsic decomposition by defining the problem in two steps. 
In the first step, we generate local and global constraints for shading discontinuities and smoothness. 
We estimate these constraints through our novel \emph{dense ordinal shading} formulation in Section~\ref{sec:ordinal}. 
Rather than regressing the absolute shading values that satisfy the intrinsic model, we define a relaxed loss that enforces the correct ordering of shading values. 
When we generate a result at the resolution of the receptive field, our ordinal shading network can generate a coherent structure for the entire scene. 
At much higher resolutions, our ordinal shading network loses global coherence due to the receptive field size being smaller than the estimation resolution. 
However, as shown in Section~\ref{sec:ordinal:highres}, it also predicts highly detailed local shading discontinuities. 
This is due to our relaxed formulation that does not regress continuous shading values but instead promotes the correct ordering of pixels. 

The low-resolution and high-resolution ordinal estimations, as a result, represent two important sets of clues about shading: the global ordering of the shading values in the entire image, and detailed discontinuities in the local neighborhood of a pixel. 
We feed these two estimations together with the image to our second network that generates our full result at high resolutions in Section~\ref{sec:metric}. 
This second network is able to generate consistent results beyond its receptive field as the global structure of the shading is provided in the form of our low-resolution ordinal input. 
It also generates highly detailed shading discontinuities thanks to the provided high-resolution ordinal input. 
Figure~\ref{fig:pipeline} shows our full pipeline. 

\begin{figure*}
    \centering
    \includegraphics[width=\linewidth]{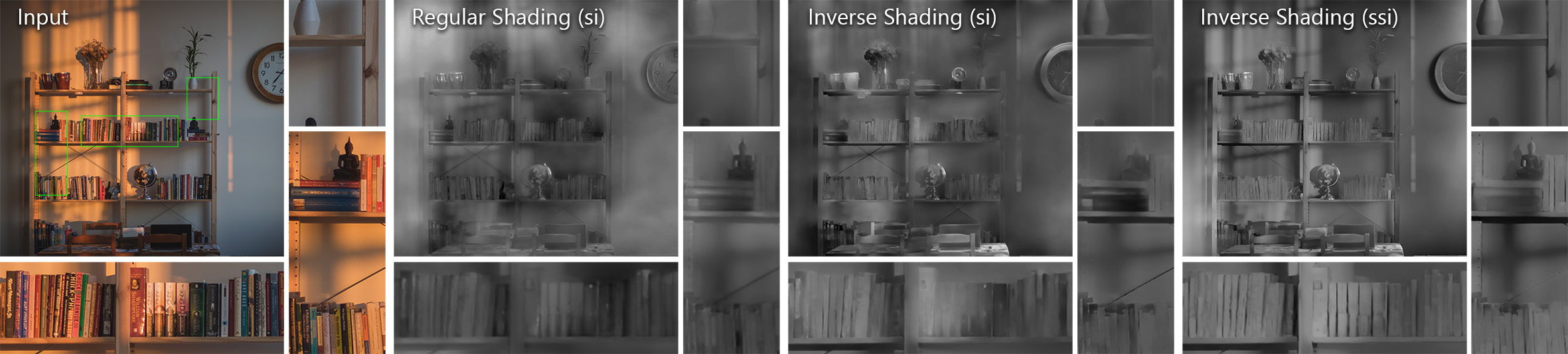}
    \caption{
    In the ablation study presented in     Section~\ref{sec:ablation:ordinal}, 
    we train the same network using the scale-invariant loss on shading, a scale-invariant loss on inverse shading, and our proposed scale-and-shift invariant ordinal formulation. 
    We show that the network trained for direct shading estimation generates very blurry results, while our ordinal formulation generates a stable and highly detailed result. 
    This confirms the idea that by solving the easier problem of ordinal shading estimation in our first step, we make better use of the limited network capacity and generate higher-quality results. 
    \hfill \footnotesize{Image from Unsplash by Lesly Juarez.}
    }
    \label{fig:method:ordinalityComparison}
\end{figure*}

In structuring our two-step setup, we took inspiration from existing literature on ordinal shading. 
In the works by \citet{zoran2015learning} and \citet{zhou2015learning}, the authors utilize CNNs to generate reflectance relationships between a sparse set of pixel pairs. 
Both works point to the easier nature of predicting ordinal relationships when compared to direct regression of the continuous values. 
Similarly, we observe an increase in the estimation quality when we use our ordinal formulation in the first step, as Figure~\ref{fig:method:ordinalityComparison} demonstrates. 
Similar to our dual resolution input to our second network, both \citet{zoran2015learning} and \citet{zhou2015learning} generate local and global ordinal relationships using the full image and local image patches. 
While we feed our \emph{dense} ordinal maps to a second network to generate the full decomposition, these methods utilize CRF-based formulations in their second step that takes the predicted sparse ordinal relationships as input. 

Our dual-resolution ordinal estimation approach is also inspired by the work of \citet{miangoleh2021boosting} that \revision{introduces} a method for boosting the resolution of pre-trained monocular depth networks. 
Similar to our work, they also generate two ordinal depth estimations at low and high resolutions and use them to generate a single high-resolution consistent estimation. 
Their second network, the \emph{merging} network, implements a low-level gradient transfer method, similar to Poisson blending~\cite{poisson}, between the two depth estimates, without the original image as input. 
On the contrary, our second network performs full intrinsic decomposition that satisfies the intrinsic equation using the ordinal estimations as constraints together with the original image as input, similar to the CRF-based optimization by \citet{zoran2015learning}.

\subsection{Inverse Shading Representation}
\label{sec:inverseShading}
 
Most natural scenes contain specular objects and objects with very dark albedos, both of which result in very large shading values. 
As a result, the shading values in a scene span a very wide range of values with long-tailed distributions. 
This skewed distribution characteristic decreases the contrast in the shading image by concentrating valid shading values in a small window inside the range. 
This makes direct regression of shading challenging especially when coupled with the scale-invariant nature of the problem. 
Some methods opt to model the problem in logarithmic domain~\cite{li2018cgintrinsics, li2018learning}. 
While the logarithmic representation increases the contrast in the distribution, it still lacks a well-defined range.

We formulate our method in both ordinal and full shading estimation in a novel \emph{inverse shading domain} defined in $[0,1]$:
\begin{equation}
    D = \frac{1}{S+1},
    \label{eq:method:inverseShading}
\end{equation}
where $S$ represents the linear-scale shading. 
The inverse shading domain creates a more uniform distribution in the $[0,1]$ range as Figure~\ref{fig:ordinal:shd_rep} demonstrates. 
This uniform distribution with high contrast is able to more accurately represent intricate shading variations such as smooth gradients and very high shading values on specular objects. 
The bounded nature of this representation is amenable to deep networks as it allows for a predictable input and output range. 
It also makes it possible to use a saturating activation function in the neural network such as sigmoid, which we find to be more stable during training when compared to commonly utilized ReLU. Note that despite the ambiguous scale, the inverse shading \revision{representation} preserves the ordinal relationships in the shading domain, i.e. $D_i < D_j$ for $S_i > S_j$ for all pixel pairs $(i,j)$.

\section{Dense Ordinal Shading Estimation}
\label{sec:ordinal}

\begin{figure*}
    \centering
    \includegraphics[width=\linewidth]{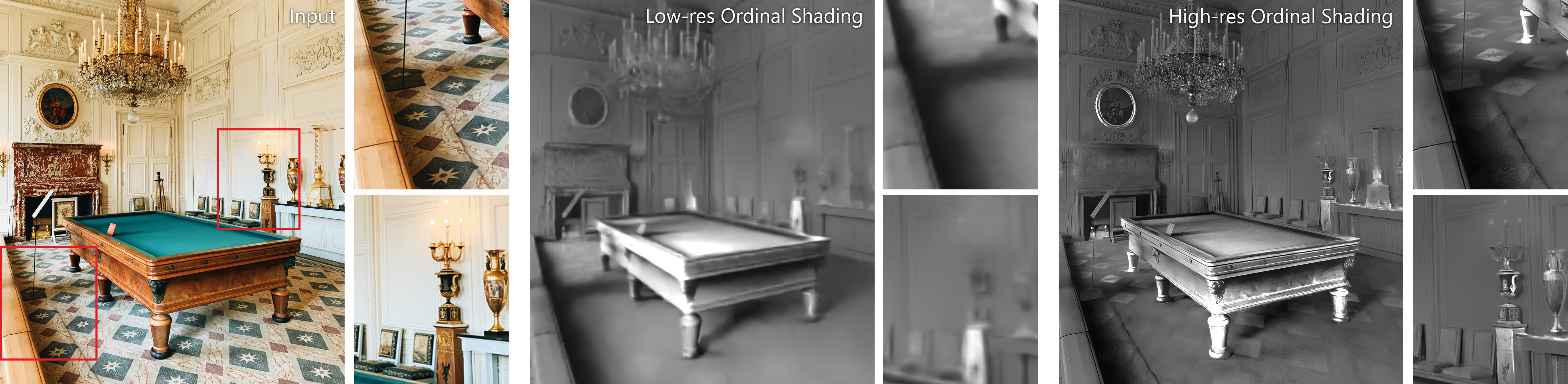}
    \caption{
    Ordinal model behavior as input resolution is increased. 
    The low-resolution estimation is computed at the receptive field size of the network and as a result, gives us a globally consistent ordinal estimation. 
    (Bottom inset) At high resolutions, on the other hand, our network generates much more detailed estimations with sharp shading discontinuities.
    (Top inset) Since we are operating at a much larger than the receptive field size, however, the high-resolution estimate may lack global coherency as seen on the carpet patterns and on the fireplace. 
    \hfill \footnotesize{Image from Unsplash by Kevin Kristhian.}
    }
\label{fig:method:multiresolution}
\end{figure*}

The shading layer in \revision{Equation}~\ref{eq:intrinsicmodel} is a continuous-valued scale-invariant variable that is required to satisfy the core model in every pixel. 
Estimating this under-constrained variable for complex scenes at high resolutions is, hence, very challenging. 
In the first step of our method, instead of estimating the continuous shading values directly, we focus on the \emph{ordinal} definition of the problem.

The ordinal definition of the problem can be seen as a relaxed reformulation of direct shading estimation~\cite{zoran2015learning}. 
Instead of satisfying the intrinsic model, the network is now only tasked with making sure the inequalities between the shading of individual pixels are satisfied. 
As a result, in the ordinal world, $f(S)$ is as valid a result as $S$ for any monotonically increasing function $f(\cdot)$, as it preserves the inequality relations in shading. %

\paragraph{Dense Ordinal Loss}
We start our dense ordinal shading formulation from this premise. 
As outlined in Section~\ref{sec:inverseShading}, we define our output in the inverse shading domain $D \in [0,1]$, in which the ordinality in shading is preserved. 
We then formulate a relaxed loss function that still ensures the correct ordering in the estimation: 
\begin{equation}
\label{eq:ordinalLoss}
    \mathcal{L}_{ord} = \frac{1}{N} \sum_{i}^{N} (f(O_i) - D^*_i)^2,
\end{equation}
where $O$ is the estimated ordinal shading, $D^*$ is the ground-truth inverse shading, and $f(\cdot)$ is a monotonically-increasing affine function defined as:
\begin{subequations}
\label{eq:axplusb}
\begin{gather}
    f(x) = ax + b \\ 
    (a, b) = \argmin_{a, b} \sum_i (f(O_i) - D^*_i)^2, \quad a>0.
\end{gather}
\end{subequations}
This way, we do not penalize the estimated result by the network if the linear ordering of the estimated shading values is accurate. 
In the end, the estimated ordinal shading $O$ is an unknown scale and shift away from the ground truth inverse shading, while satisfying the ground-truth ordinal relationships.

The network trained with our dense ordinal loss learns to correctly order the shading values for each pixel, while not necessarily satisfying the intrinsic model. 
This simplified problem definition allows the network to generate ordinal results with more high-resolution details when compared to direct shading estimation as Figure~\ref{fig:method:ordinalityComparison} demonstrates. 

Our dense ordinal loss formulation has some similarities to the scale- and shift-invariant loss formulation by \citet{ranftl2020towards} for monocular depth estimation. 
In their formulation, the affine loss allows them to use the stereo depth datasets together with metric depth datasets.
This is due to the disparities computed from stereo data being accurate up to a scale and shift from the metric ground truth. 
In our case, however, the scale and shift loss is applied solely to promote ordinality. 
While it can be replaced with any monotonically-increasing function in our formulation, we utilize this first-order function because of its stability during training.

\revision{Note that our dense ordinal shading definition is a generalization of the pair-wise ordinality previously explored in the literature~\cite{zoran2015learning,narihira2015learning,zhou2015learning}. 
As Equations~\ref{eq:ordinalLoss} and \ref{eq:axplusb} imply, we estimate a dense result that has the same resolution as the input image that maintains the pair-wise ordinality between the shading values of any pixel pair thanks to the monotonically increasing transformation $f(\cdot)$. 
}

\paragraph{Smoothness Loss}
In addition to the ordinal affine loss $\mathcal{L}_{ord}$, we also use the multi-scale gradient loss $\mathcal{L}_{msg}^o$~\cite{li2018mega, li2018cgintrinsics} on the ordinal shading:
\begin{equation}
    \mathcal{L}_{msg}^o = \frac{1}{NM} \sum_{i=1}^N \sum_{l=1}^M |\nabla f(O)_{i,l} - \nabla D^*_{i,l}|
    \label{eq:method:ord_grad_loss}
\end{equation}
where $\nabla$ denotes the spatial gradient and $l$ denotes the level of a multi-scale pyramid. 
We found this loss defined on the gradients to be essential in generating spatially consistent estimations. 

Our overall loss function for ordinal shading estimation is: 
\begin{equation}
    \mathcal{L}_{os} = \mathcal{L}_{ord} + \lambda_{msg}^o \mathcal{L}_{msg}^o
\end{equation}
where we set $\lambda_{msg}^o$ as $0.5$. 
This loss is evaluated on the single-channel ordinal shading estimation generated by the network.

\paragraph{Network Architecture}
We employ the architecture defined in~\cite{ranftl2020towards} which consists of the encoder-decoder network proposed in~\cite{lin2017refinenet} with a ResNext101 encoder from~\cite{xie2017aggregated}. We do not use any pre-trained weights and instead train the network from scratch. We add a sigmoid activation to the end of the network to bound our predictions to $[0, 1]$. We train the network using the Adam optimizer with a learning rate of $10^{-5}$.

We train our ordinal shading network using a variety of synthetic datasets as well as the dense real-world dataset we derive from multi-illumination data~\cite{murmann19multi}. 
We go into the details of how we combine synthetic and real-world data in Section~\ref{sec:datasets}.

\subsection{Multi-Resolution Behavior}
\label{sec:ordinal:highres}

Given its fully-convolutional nature, our ordinal shading network is able to generate ordinal predictions at resolutions beyond the training resolution of $384 \times 384$. 
While the network is able to generate high-resolution outputs, the input resolution changes the characteristics of the estimated ordinal shading.

When the input image is resized to the training resolution for inference, as many data-driven setups do by default~\cite{luo2020niid, das2022pie, liu2020unsupervised}, we can generate a consistent shading structure for the entire scene. 
In this scenario, since the entire image fits in the receptive field size of the network, we see a consistent shading structure in the estimation. 
This is due to how convolutional neural networks operate: since the network can \emph{see} the entire image at once, it is able to generate global ordinal shading constraints. 
However, especially in complex scenes, many high-resolution details are missing in this estimation as Figure~\ref{fig:method:multiresolution} shows. 
The lack of high-resolution details comes from the limited capacity of the network: a network that has to produce a complex shading structure can not also generate high-resolution details. 

When the input image is fed to the network at higher resolutions than its receptive field, we see in Figure~\ref{fig:method:multiresolution} that we can generate intricate shading variations in high resolution. 
However, this comes at the cost of global coherency. 
This is a result of the limited receptive field size where the network can not produce the correct ordering of pixels that are spatially far away from each other. 
However, as the receptive field sees image patches with \revision{lower} local scene complexity, it can now generate high-resolution details much more accurately. 
This points to the fact that there is a trade-off in ordinal shading estimation quality with increasing input resolution. 
While the low-resolution estimations create reliable global constraints, the high-resolution estimations can generate detailed shading discontinuities. 

A similar observation was made by \citet{miangoleh2021boosting} for ordinal depth estimation networks. 
This parallel between the two tasks is not surprising, as monocular depth estimation and shading estimation share many similarities especially in the ordinal definition of the problems as \citet{zoran2015learning} points out.
\citet{miangoleh2021boosting}, in their analysis of monocular depth estimation at different resolutions, proposes to use the image edge density to determine the resolution at which the network can still produce consistent results. 
This resolution, defined as $\mathcal{R}_0$, is computed as the largest resolution in which every receptive field-sized region in the image contains strong image edges. 

We observed that $\mathcal{R}_0$, which depends on image content, is a good upper limit for ordinal shading estimation to produce high-resolution details while creating spatially coherent results. 
Given the limited capacity of our networks and lack of very high-resolution training data, we also limit the size of input images to have at most 1500 pixels in either width or height.
We find that using these resolutions for ordinal shading estimation provides the intrinsic decomposition network with reliable local ordinal constraints with detailed high-frequency shading discontinuities.

\section{Intrinsic Decomposition with Ordinal Input}
\label{sec:metric}

We set up the high-resolution intrinsic decomposition problem with two ordinal shading estimations provided as input to the network together with the linear RGB image. 
The first ordinal estimation, $O_L$, is generated at the receptive field size of our ordinal shading network, $384 \times 384$. 
Since it is generated at the receptive field resolution, the ordinal network is able to generate an accurate ordering of the shading values across the entire image, using all the information present in the scene. 
$O_L$, in the end, provides a reliable general structure for the final shading estimation and provides the global constraints for the entire image.
The second ordinal estimation, $O_H$, is generated at a much higher $\mathcal{R}_0$ resolution as outlined in Section~\ref{sec:ordinal:highres}. 
As this estimation is generated at a higher resolution than the receptive field, it lacks global coherency. 
However, it contains high-resolution shading discontinuities, providing intricate local constraints to our second network.
\revision{We provide the ordinal inputs $O_L$ and $O_H$ by concatenating them with the input image across the color channel dimension to create an input to our second network of size $(H \times W \times 5)$ after upscaling $O_L$ to the resolution of $O_H$. 
}
An overview of our pipeline is presented in Figure~\ref{fig:pipeline}.

This contextually rich information readily provided to the network simplifies the intrinsic decomposition task. 
Our decomposition network is not required to conduct the high-level task of inferring the overall structure of the shading, which requires reasoning about global context such as geometry and illumination direction, as it is provided in the form of $O_L$. 
At the local level, similarly, our network is not required to determine whether a strong RGB edge comes from a shading discontinuity or a sudden change in the albedo, as it is provided in the form of $O_H$. 
As a result, the task of our decomposition network becomes adjusting the overall structure from $O_L$ to satisfy the intrinsic model in Equation~\ref{eq:intrinsicmodel} while integrating the details in $O_H$ to the final estimation. 
This way, we are able to regress continuous shading values with intricate details at high resolutions.

\begin{figure}
    \centering
    \includegraphics[width=\linewidth]{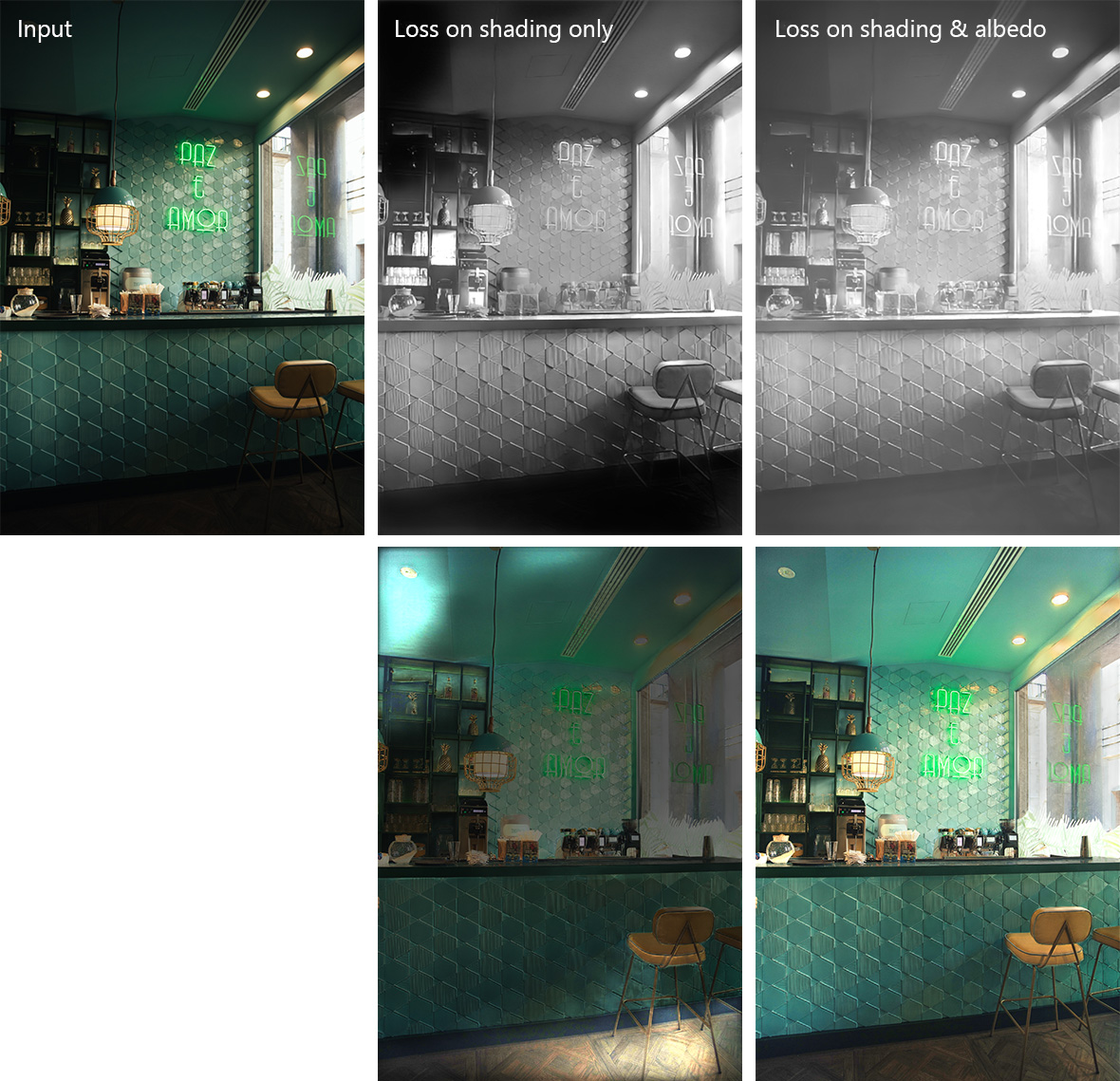}
    \caption{
    Regions with very low shading values are particularly challenging due to the weak color signal present in the input. 
    When we only define our losses on the shading, due to the division operation of two small numbers when we compute our albedo using Equation~\ref{eq:intrinsicmodel}, artifacts may appear in these regions. 
    Once we add the losses on albedo in the training of the same system, the network is able to create stable estimations with sparse and accurate albedo.
    \hfill \footnotesize{Image from Unsplash by Logan Stone.} 
    }
    \label{fig:highdef_loss_comp}
\end{figure}

\subsection{Output Formulation}

Our intrinsic decomposition network generates the result in the single-channel inverse shading domain. 
We then compute the shading using Equation~\ref{eq:method:inverseShading} as well as the albedo using the intrinsic model in Equation~\ref{eq:intrinsicmodel}:
\begin{equation}
S = \frac{1 - D}{D}, \quad\quad A = \frac{I}{S} = \frac{I*D}{1-D},
\label{eq:method:finaloutputs}
\end{equation}
where $I$ and $D$ represent the input image and the estimated inverse shading, respectively. 

Most intrinsic decomposition methods~\cite{narihira2015direct, shi2016learning, baslamisli2018cnn, cheng2018intrinsic, li2018cgintrinsics, li2018learning, zhou2019glosh, luo2020niid, das2022pie} estimate the shading and albedo as two different variables and formulate a reconstruction loss using Equation~\ref{eq:method:inverseShading}. 
This often results in the estimated $S$ and $A$ not being able to reconstruct the image $I$ correctly. 
In contrast, we follow the works by \citet{fan2018revisiting} and  \citet{lettry2018darn,lettry2018unsupervised} and compute the albedo from the estimated shading, making sure the intrinsic model in Equation~\ref{eq:intrinsicmodel} holds. 

Inferring the albedo from the estimated inverse shading also allows us to define independent losses on the two components. 
While they are related through the input image, the albedo and shading losses have complementary information to each other. 
As detailed in the rest of this section, we define losses on the albedo in addition to the inverse shading and backpropagate them through Equation~\ref{eq:method:finaloutputs}. 
This way, we can generate high-resolution shading estimations that yield accurate albedo components at the same time as Figure \ref{fig:highdef_loss_comp} demonstrates.

\begin{figure*}
    \centering
    \includegraphics[width=\linewidth]{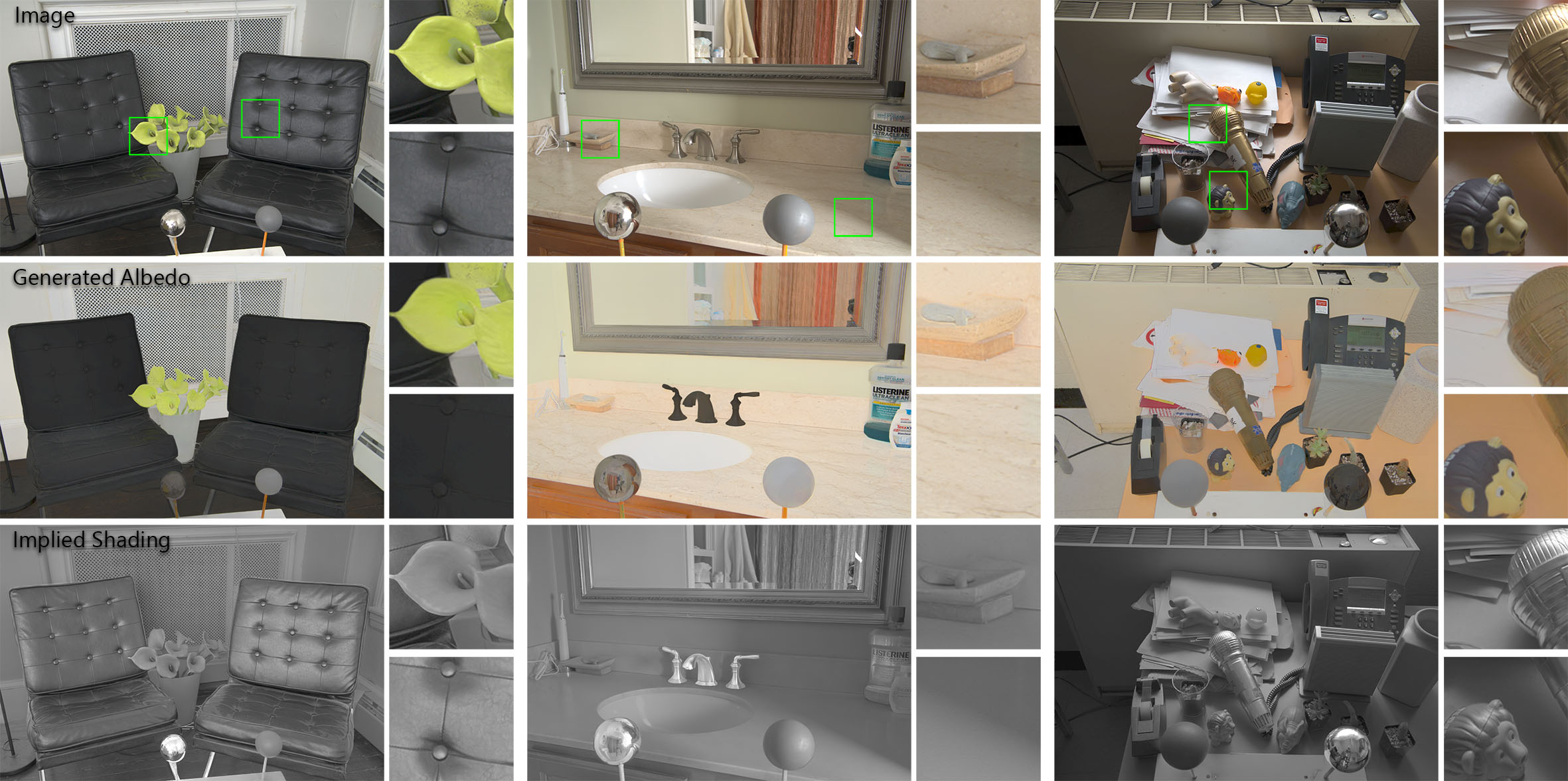}
    \caption{
    Examples of the pseudo-ground truth albedo and shading pairs we generate from images in the Multi-Illumination Dataset \cite{murmann19multi}}
    \label{fig:pseudo_gt_examples}
\end{figure*}

\subsection{Scale Ambiguity}
Given the inherent scale ambiguity of the intrinsic decomposition formulation, prior approaches rely on scale-invariant losses to supervise deep networks. 
These scale-invariant losses require a least-squares fit between the network estimation and the ground truth during training. 
At later stages of the training, this scale can be computed effectively using least squares. 
However, during the initial stages of training, the scale must be computed between the ground truth shading and the highly inaccurate estimations from the under-trained network due to the outlier-prone nature of least-squares.

In our setup with ordinal inputs, our globally consistent low-resolution ordinal estimation $O_L$ already provides a point of reference to our intrinsic decomposition network. 
Since we train our ordinal network before the intrinsic network and compute $O_L$ using this well-trained system, the least-squares fit between $O_L$ and the ground-truth yields stable results.
Hence, we use our low-resolution input to set the arbitrary scale in the ground truth:
\begin{equation}
    c = \argmin_x \sum_i (x A^{**} - \tilde{A}_L)^2, \quad\quad \tilde{A}_L = \frac{I}{\tilde{S}_L}, \quad\quad \tilde{S}_L = \frac{1 - O_L}{O_L}
\end{equation}
where $A^{**}$ represents the ground-truth albedo at an arbitrary scale. 
We then use this fixed scale to define our ground-truth shading, inverse shading, and albedo:
\begin{equation}
    A^* = c * A^{**}, \quad \quad S^* = \frac{I}{A^*}, \quad \quad D^* = \frac{1}{S^* + 1}.
\end{equation}
We perform the scale matching on the albedo component due to its limited range of values unlike shading with long-tailed distributions. We similarly fix the average scale of the high-resolution ordinal input $O_H$ to that of $O_L$ to get the input and output variables at the same overall scale. By fixing the arbitrary scale, we can define dense losses without the need for scale invariance. We provide additional discussion on scale-invariant losses in Section D.1 of the supplementary material.

\begin{figure*}
    \centering
    \includegraphics[width=\linewidth]{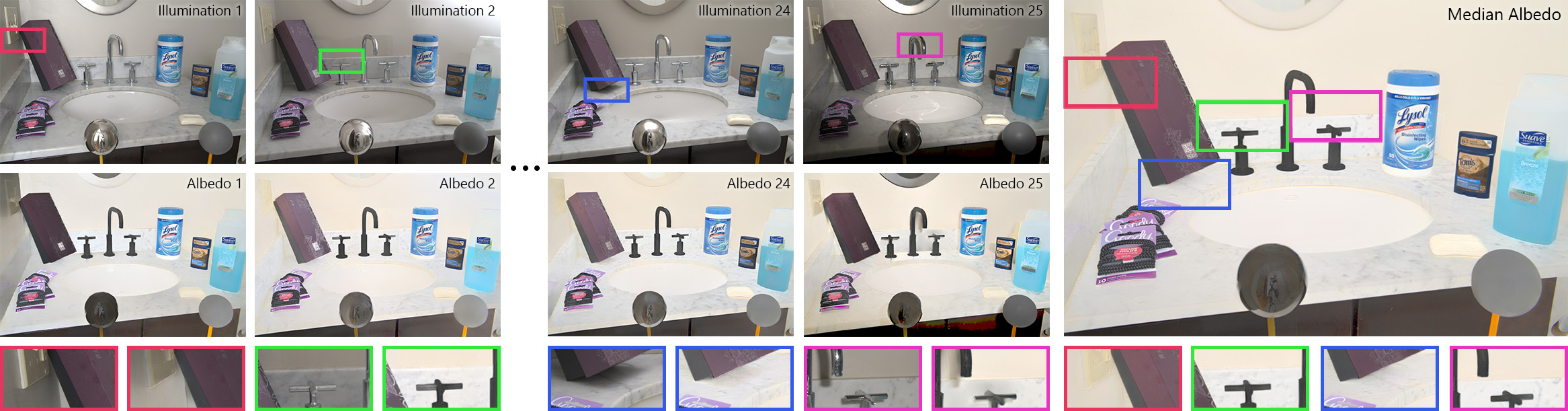}
    \caption{
    Each scene in the Multi-Illumination Dataset \cite{murmann19multi} is captured under 25 different illuminations. 
    These sets of photographs provide us with the observation of the same albedo combined with different shadings. 
    After our initial synthetic-only training, we estimate the albedo for all 25 illuminations. 
    Each estimated albedo may contain imperfections especially in challenging regions such as strong shadows or specular objects as the insets show. 
    We compute our pseudo-ground truth albedo for each scene by taking the per-pixel median of our individual albedo estimations. 
    Due to the abundance of data, we show that we can generate a robust and reliable estimation of the shared albedo map, which we then use to further train our network. 
     }
    \label{fig:multi_illum_pipeline}
\end{figure*}
\subsection{Dense Losses}

We use an L1 loss together with the multi-scale gradient loss on the estimated inverse shading:
\begin{subequations}
    \begin{gather}
        \mathcal{L}_{s} = \frac{1}{N} \sum_{i=1}^N |D_i - D^*_i| \\
        \mathcal{L}_{msg}^s = \frac{1}{NM} \sum_{i=1}^N \sum_{l=1}^M |\nabla D_i - \nabla D^*_{i,l}|,
    \end{gather}
\end{subequations}
where $\nabla$ denotes the spatial gradient, and $l$ denotes the level of a multi-scale pyramid. 
Our choice of defining the loss on the inverse shading instead of shading comes from the constrained range of inverse shading between $[0,1]$. 
The very large shading values in $S$, coming from the long-tailed distribution shown in Figure~\ref{fig:ordinal:shd_rep}, results in skewed values in the loss. %
The balanced distribution of shading values in $[0,1]$ also helps to better represent shading gradients and intricate details around specular regions.

We also define the same two losses on albedo:
\begin{subequations}
    \begin{gather}
        \mathcal{L}_{a} = \frac{1}{N} \sum_{i=1}^N |A_i - A^*_i| \\
        \mathcal{L}_{msg}^a = \frac{1}{NM} \sum_{i=1}^N \sum_{l=1}^M |\nabla A_i - \nabla A^*_{i,l}|,
    \end{gather}
\end{subequations}
and backpropagate these through Equation~\ref{eq:method:finaloutputs} to combine them with our losses on the inverse shading. 

By defining our output as the inverse shading and deriving the estimated albedo using the intrinsic \revision{model}, we are able to define and combine losses on both albedo and shading components. 
Albedo and shading maps have their own unique characteristics that are helpful for the network to model the underlying statistics. 
For instance, the smoothness of shading represents continuous geometric surfaces, while high gradients in the shading represent geometric discontinuities or shadows. The high-gradient regions on the albedo, however, represent textured surfaces while we expect a smooth albedo across shadows. 
By defining the multi-scale gradient loss on both components, we are able to signal both of these characteristics to the network at the same time, making both our shading and albedo estimates more accurate. 
While the joint loss overall improves our accuracy as analyzed in Section~\ref{sec:ablation:jointloss}, it significantly improves our results in images with very dark regions as shown in Figure~\ref{fig:highdef_loss_comp} by making use of the albedo priors when there is not enough signal in the regions with very dark shading.

\subsection{Network Structure and Training}
We utilize the same encoder-decoder architecture used for our ordinal network. Given the simplified nature of this task, we opt for a smaller and more memory-efficient EfficientNet \cite{tan2019efficient} encoder. We add a sigmoid activation to the end of the network to bound our inverse shading estimation to $[0, 1]$. We train the network using the Adam optimizer with a learning rate of $10^{-5}$.

We omit the low-resolution rendered datasets and train the network on a mixture of images from the Hypersim Dataset \cite{roberts2021hypersim} and the MID-Intrinsics dataset we derive from the Multi-Illumination Dataset \cite{murmann19multi}. We provide details on the preparation and use of these datasets in Section~\ref{sec:datasets}.

\section{Training on Rendered and Real-World Data}
\label{sec:datasets}

It is notoriously hard for intrinsic decomposition networks to generalize to real-world images from synthetic-only training data~\cite{garces2022survey}. 
It is also challenging to create dense ground-truth data for real images. 
Due to this difficulty, the only real-world ground truth datasets available are in the form of sparse ordinal comparisons~\cite{bell2014intrinsic,kovacs17shading}. 
These sparse annotations have been very useful as the only form of real-world supervision. 
However, they do not allow the use of any dense loss on shading, albedo, or their gradients. 
This limits their usefulness in generalizing to in-the-wild input.

In this work, we derive a dense, high-resolution real-world intrinsic decomposition dataset by exploiting the redundant information present in photographs of the same scene captured under different illuminations. 
We first train our networks with a set of synthetic datasets. 
Then, using our network, we formulate a robust albedo estimator that estimates a single consistent albedo using all 25 illuminations for each of 1000 scenes presented in the Multi-Illumination Dataset by Murmann et al.~\shortcite{murmann19multi}. 
We show that using this indoor real-world dense dataset allows our system to generalize to in-the-wild examples with challenging phenomena such as out-of-focus blur, as well as \revision{outdoors photographs} and novel subjects not present in the training datasets such as the human face as Figure~\ref{fig:multi_comp} demonstrates. \revision{In the supplementary material, we provide implementation details pertaining to the training of our pipeline and the preprocessing performed for each dataset.}

\begin{figure*}
    \centering
    \includegraphics[width=\linewidth]{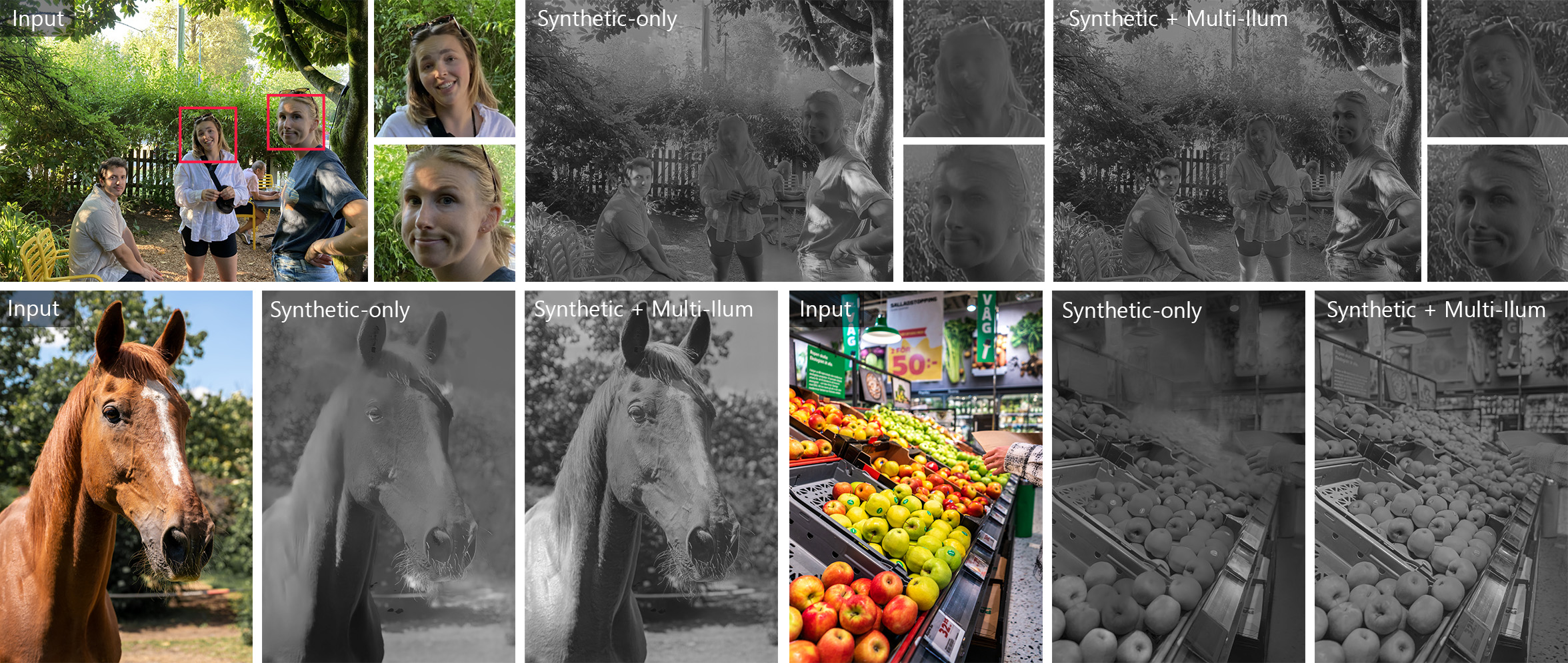}
    \caption{
    Although the Multi-Illumination Dataset only contains indoor scenes, combining this real-world dataset with synthetic data during training is beneficial for generalizing to in-the-wild photographs. 
    Our system generalizes to previously unseen objects such as the human face and animals and can generate consistent results in regions with out-of-focus blur. 
    \hfill \footnotesize{Images from Unsplash by Ladislav Stercell (horse) and Philip Myrtorp (apples).}
    }
    \label{fig:multi_comp}
\end{figure*}

\paragraph{Synthetic Training}

We first start our training with the ordinal shading network, as the decomposition network requires ordinal estimations as input. 
We make use of the synthetic datasets CGIntrinsics \cite{li2018cgintrinsics}, OpenRooms \cite{li2021openrooms}, and Hypersim \cite{roberts2021hypersim} that are composed of realistic renderings of indoor environments.
We also use the GTA Dataset \cite{krahenbuhl2018free} which includes outdoor environments from the video game \emph{Grand Theft Auto V}. 
We train our ordinal network for 700,000 iterations with a batch size of $8$, sampling images from each dataset.%

We then generate the low-resolution and high-resolution ordinal inputs for images in the Hypersim dataset \cite{roberts2021hypersim} and train our decomposition network for 200,000 iterations also with a batch size of 8. 
We only use Hypersim to train this second network as a synthetic dataset, as it is the only high-resolution dataset that allows us to train at the resolution of $(512 \times 512)$.
At this point, our system is effective in scenes that are similar to the training data such as indoor environments present in the Multi-Illumination Dataset \cite{murmann19multi}.

\paragraph{Redundant Albedo in Multiple Illuminations}
The Multiple Illumination Dataset (MID) presents each scene under 25 different ambient lighting conditions obtained through a rotating motorized flashlight with a static camera. 
With the shared scene content and varying illumination, these 25 photographs per scene have the same albedo with different shadings. 
\revision{We propose to estimate this shared albedo using our shading predictions for the 25 photographs for each scene.} 

At this point in our training, our system can estimate a good albedo for each photograph. 
However, each individual estimation is far from perfect especially in challenging regions such as poorly illuminated parts of the scene, shadow boundaries, or specularities. 
Due to the varying illumination across photographs, these imperfections appear independently in different parts of each image as shown in Figure~\ref{fig:multi_illum_pipeline}. 

We estimate the shared albedo robustly by combining our 25 individual estimations of the same shared albedo. 
We start by matching the scale of our individual albedo estimations. \revision{We arbitrarily choose the first estimated albedo in each set to determine the scale of the rest of the estimations}:
\begin{equation}
    \tilde{A}_k = \left(\argmin_x \sum_i (\tilde{A}_1 - x\tilde{A}_k^*)^2\right) \tilde{A}_k^*, \quad k \in \{2, 3, ..., 25\},
\end{equation}
where $\tilde{A}_k^*$ and $\tilde{A}_k$ represent the initial and scale-adjusted albedo estimations for the $k^{\text{th}}$ photograph, respectively, \revision{and the summation is defined over every pixel $i$   }. 
Once all the individual albedo estimations are at the same scale, we compute the robust shared albedo component as the per-pixel median across the 25 estimations:
\begin{equation}
    A^{**} = \text{median}\left( \{\tilde{A}_k\}_{k=1}^{25} \right).
\end{equation}
This median operation, due to the high number of estimations, is able to filter out many of the issues in the individual estimations as Figure~\ref{fig:multi_illum_pipeline} demonstrates. 
Figure \ref{fig:pseudo_gt_examples} shows a few examples of the resulting pseudo-ground truth generated by this process. 
Since each sequence contains $25$ images, this method can generate sparse albedo components and shading even for difficult images that contain hard shadows, flash illumination, and specularity.
We share more examples in the supplementary material. 
In the end, we are able to generate high-quality shared albedo estimations that we use to train our system further. Our intrinsic decomposition dataset derived from MID contains 25,000 unique real-world input images and their corresponding intrinsic components at 1.5-megapixel resolution. 
We will release this data publicly for further research use.

\begin{figure*}
    \centering
    \includegraphics[width=\linewidth]{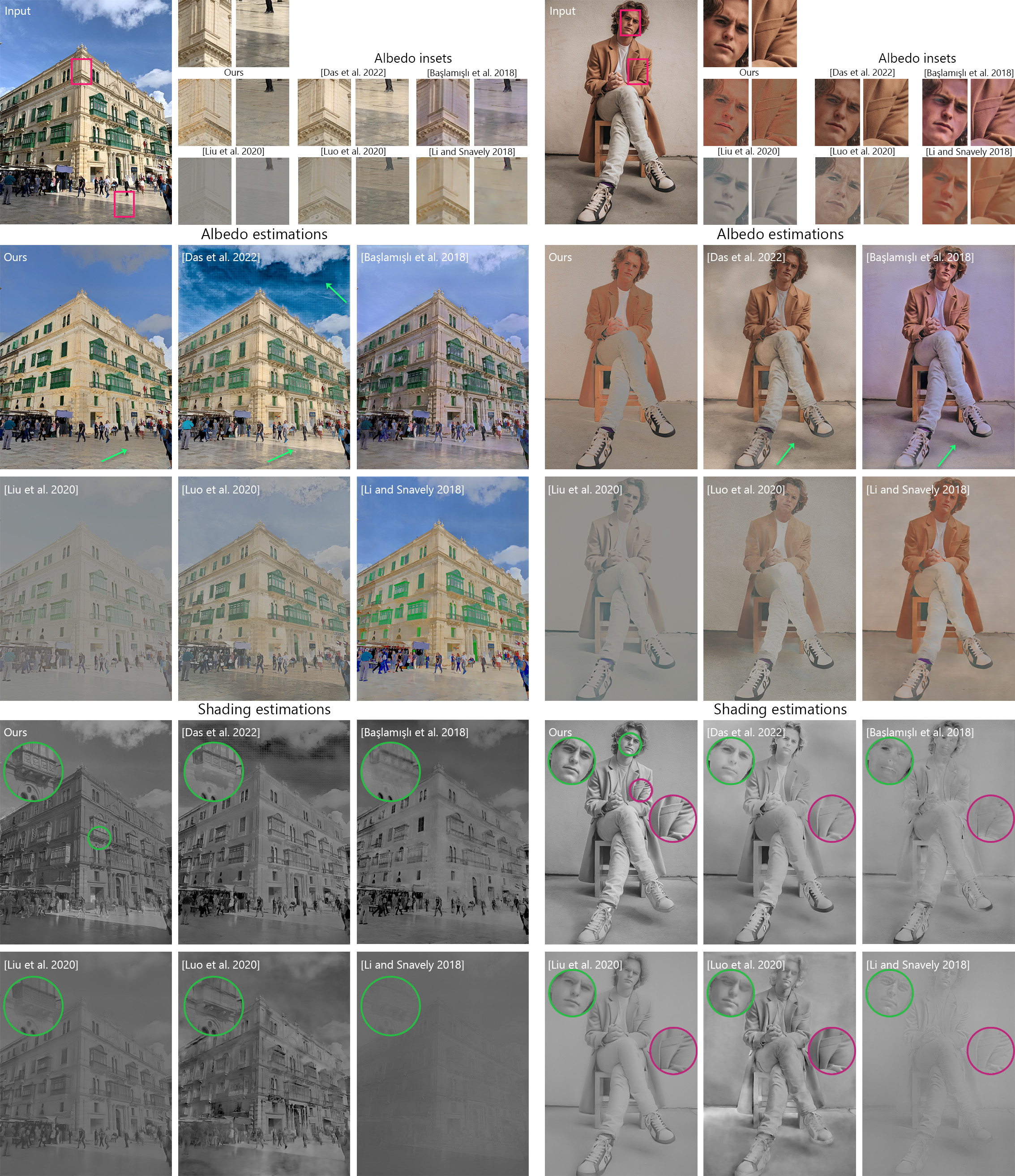}
    \caption{
    Comparison of state-of-the-art methods on in-the-wild photographs. Please refer to Section~\ref{sec:itw_comp} for a detailed discussion. 
    \\ \text{  }
    \hfill \footnotesize{Images from Unsplash by William Jones (left) and Austin Wade.}    }
    \label{fig:itw_main_comp}
\end{figure*}
\begin{figure*}
    \centering
    \includegraphics[width=\linewidth]{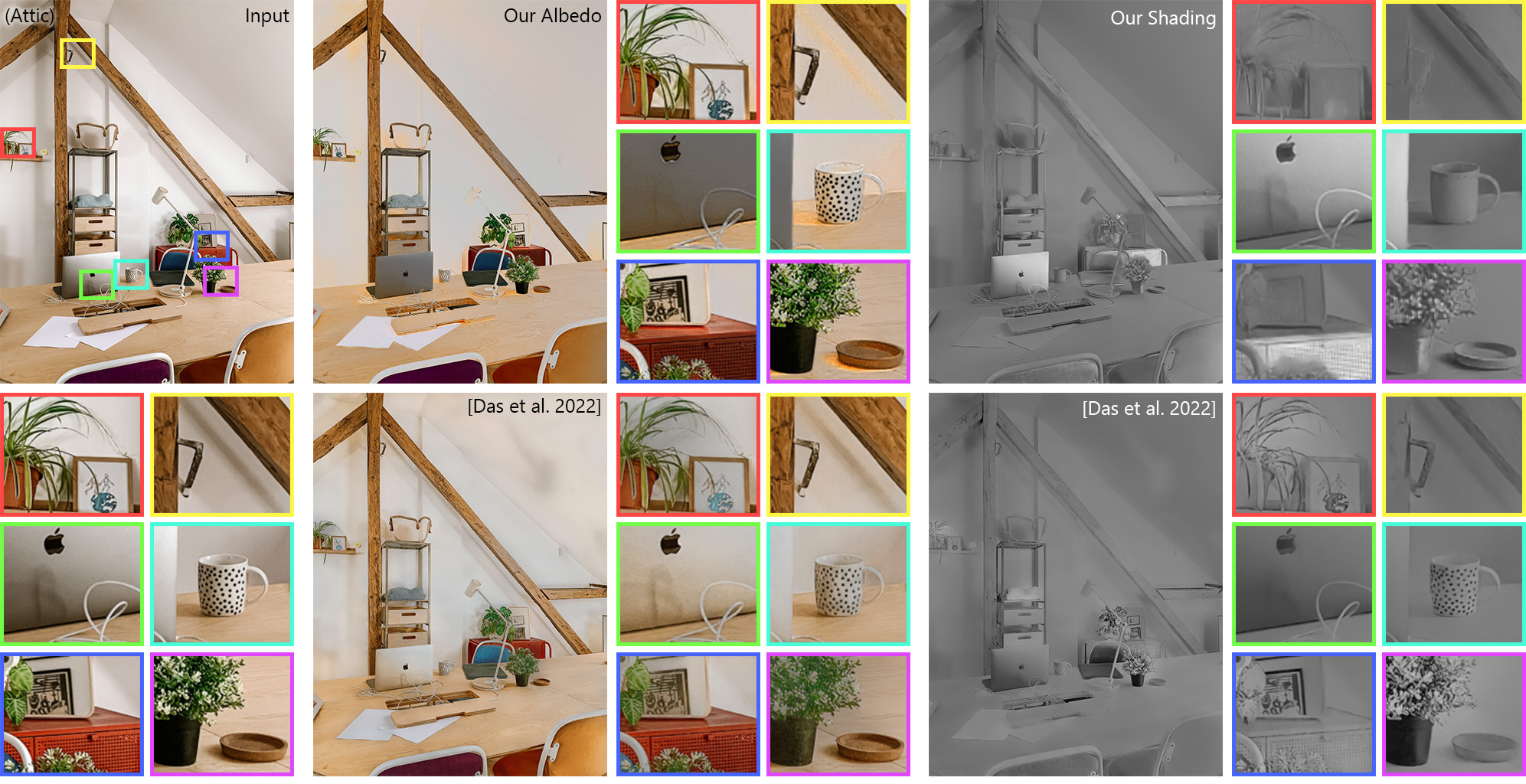}\\
    \includegraphics[width=\linewidth]{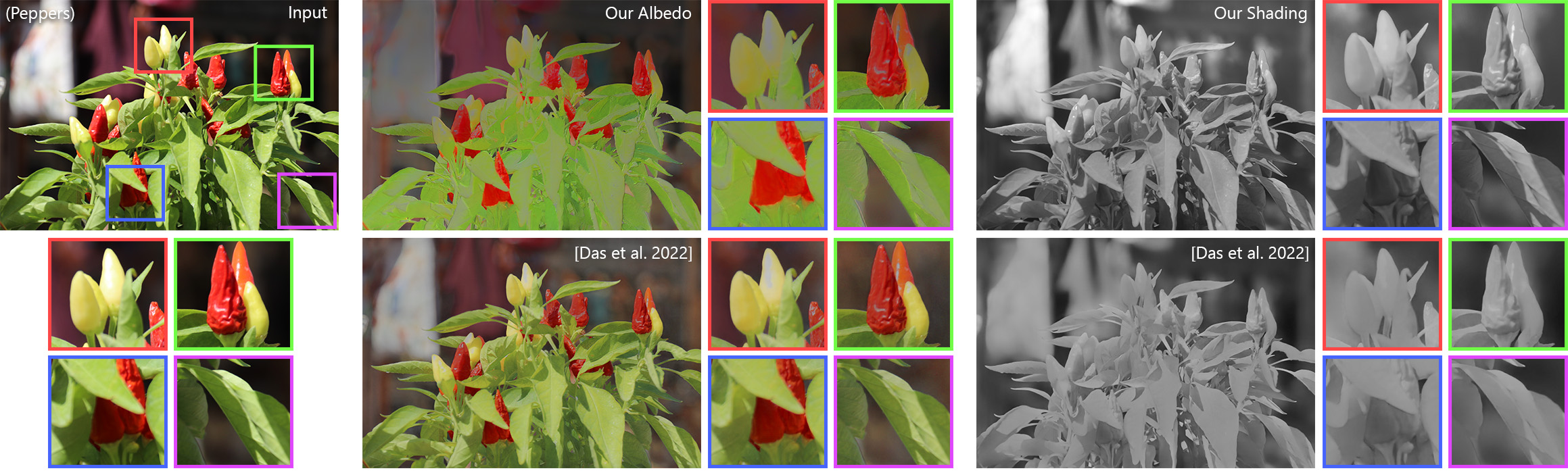}
    \caption{
    Comparison between our method and that of Das et al.~\shortcite{das2022pie} on in-the-wild photographs. Please refer to Section~\ref{sec:itw_comp} for a detailed discussion. 
    \\ \text{  }
    \hfill \footnotesize{Images from Unsplash by Toa Heftiba (Attic) and Shalev Cohen (Peppers).}
    }
    \label{fig:itw_das_comp}
\end{figure*}

\begin{figure*}
    \centering
    \includegraphics[width=\linewidth]{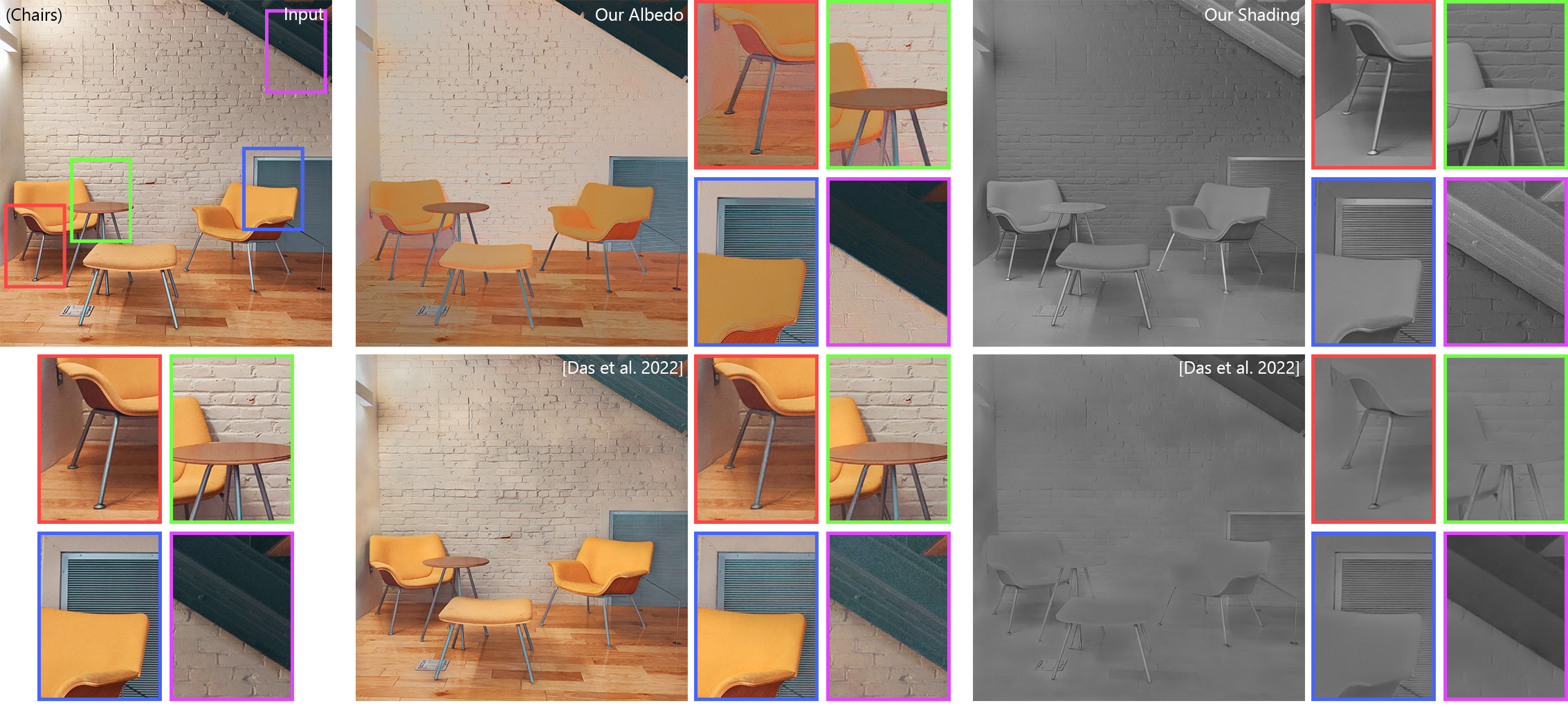}\\
    \includegraphics[width=\linewidth]{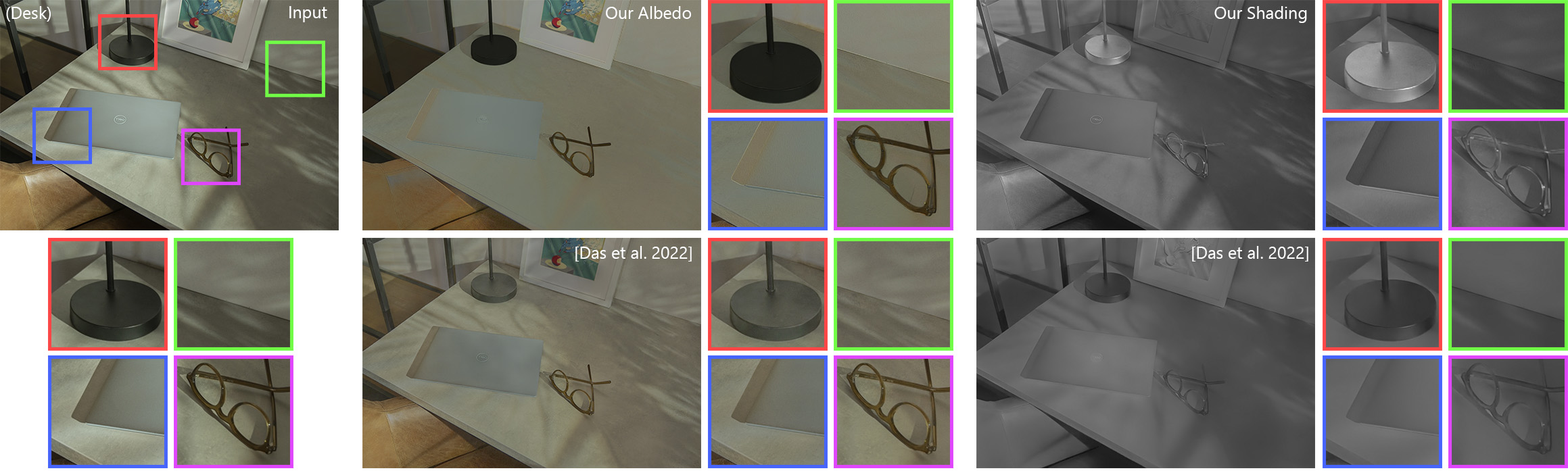}
    \caption{    
    Comparison between our method and that of Das et al.~\shortcite{das2022pie} on in-the-wild photographs. Please refer to Section~\ref{sec:itw_comp} for a detailed discussion. 
    \\ \text{  }
    \hfill \footnotesize{Images from Unsplash by Avi Waxman (Chairs) and Dell (Desk).}
    }
    \label{fig:itw_das_comp2}
\end{figure*}

\begin{figure*}
    \centering
    \includegraphics[width=\linewidth]{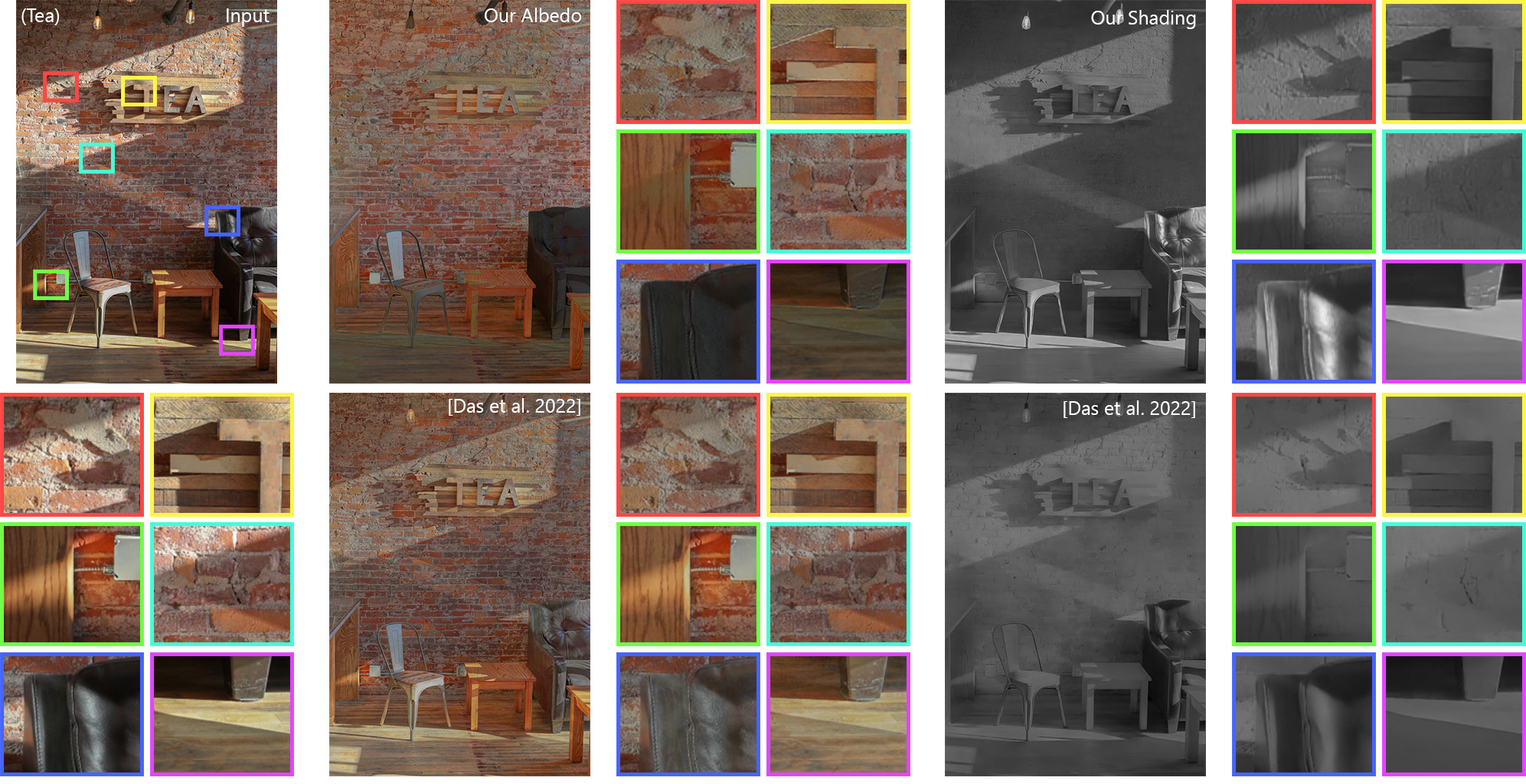}\\
    \includegraphics[width=\linewidth]{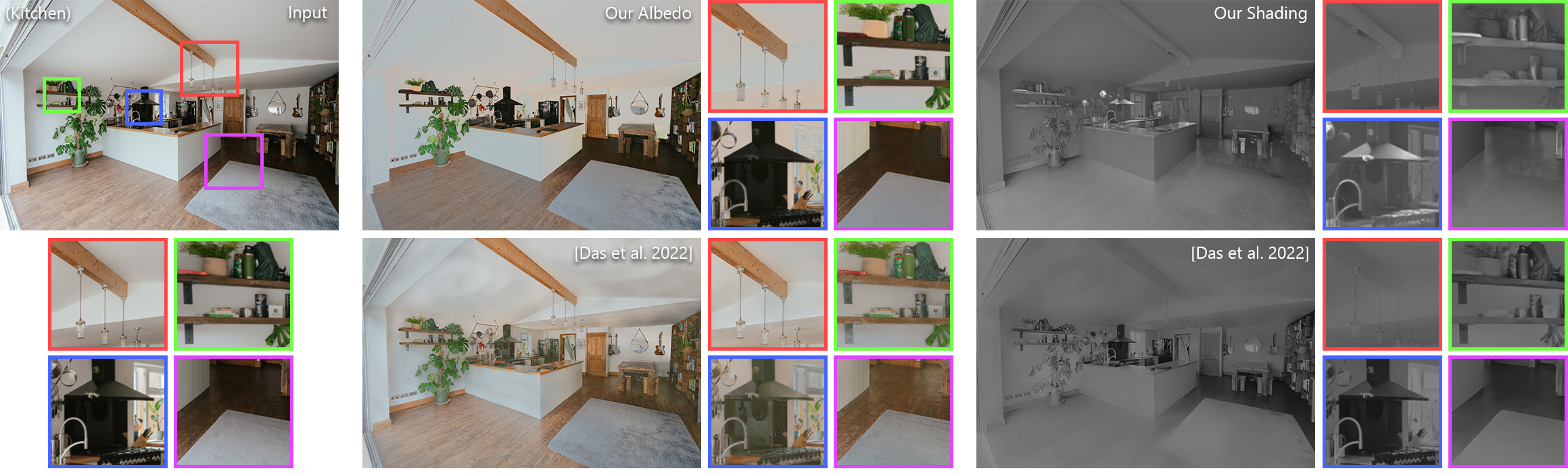}
    \caption{    
    Comparison between our method and that of Das et al.~\shortcite{das2022pie} on in-the-wild photographs. Please refer to Section~\ref{sec:itw_comp} for a detailed discussion. 
    \\ \text{  }
    \hfill \footnotesize{Images from Unsplash by Randy Fath (Tea) and Annie Spratt (Kitchen).}
    }
    \label{fig:itw_das_comp3}
\end{figure*}
\paragraph{Generalization to in-the-wild photographs}

We continue training both of our networks using the generated multi-illumination ground-truth pairs. We observe a clear improvement in model generalization after training with the multi-illumination data. Figure \ref{fig:multi_comp} shows the difference in predictions when training with and without the multi-illumination data. For fair comparison, both versions of the pipeline are trained for the same number of total iterations. The model trained with multi-illumination data is able to make accurate predictions on out-of-distribution scenes such as human faces. Additionally, we notice an increase in detail on high-frequency content such as the leaves and hair. The model is also able to make reliable predictions even in the presence of difficult camera effects such as out-of-focus blur.

\section{Qualitative Evaluation}
\label{sec:itw_comp}

Generalization to in-the-wild photographs is an important challenge that intrinsic decomposition literature is yet to address~\cite{garces2022survey}. 
We start our evaluation by comparing our performance on a variety of scenes to that of state-of-the-art methods in the literature. \revision{We focus our comparison on recent learning-based methods with open-source implementations.}
We compare our method to 
PIE-Net~\cite{das2022pie}, 
the joint semantic segmentation and intrinsic decomposition method (JSI) by \citet{baslamisli2018joint},
the unsupervised approach (USI) by \citet{liu2020unsupervised}, 
NIID-Net~\cite{luo2020niid}, 
and CGIntrinsics~\cite{li2018cgintrinsics}. 
For all of these methods, we use the publicly available implementations provided by the authors. 
We further present comparisons against the methods by \citet{lettry2018unsupervised}, \citet{bell2014intrinsic}, \citet{shen2011intrinsic}, and \citet{li2018learning} in the supplementary material.

\paragraph{Comparison of estimated albedos}
Figure~\ref{fig:itw_main_comp} shows an outdoor scene and a portrait taken in a studio environment. 
When we analyze the albedo estimations, we see that most methods have trouble generating a smooth albedo across shadow boundaries. 
PIE-Net, JSI, NIID-Net, and CGIntrinsics fail to disentangle the shadows in their estimated albedo. 
USI and NIID-Net generate albedo maps with little contrast and color content, while JSI slightly shifts the color of the scene overall. 
Similar shortcomings are also prominent in the portrait of the man, where most methods fail to remove the shadows from the man's face, on his clothing, and on the ground. 
The lack of contrast for USI and NIID-Net as well as the color shift in JSI are present in this scene as well. 
In the outdoor scene, our method is able to estimate a sparse albedo that remains consistent across the self-shadows on the building as well as the cast shadow on the ground.
Similarly, for the portrait image, we can generate a flat color for the face and the jacket, and remove the shadow on the ground from the albedo. 

\paragraph{Comparison of estimated shadings}
When we examine the shading estimations in Figure~\ref{fig:itw_main_comp}, we see the competing methods struggling with a lack of sharpness and contrast in their results. 
CGIntrinsics \cite{li2018cgintrinsics} and USI \cite{liu2020unsupervised} are generally prone to generating very smooth shading maps that lack sharp shading discontinuities. 
NIID-Net \cite{luo2020niid}, on the other hand, generates low-frequency artifacts, especially in high-resolution estimations. 
PIE-Net \cite{das2022pie} and JSI \cite{baslamisli2018joint} often fail to reflect the shadow boundaries in shading, which are incorrectly represented in the corresponding albedo estimations. 
Our method is able to successfully represent the sharp shading discontinuities across shadow boundaries and generate realistic, high-contrast shading maps. 
It can be seen on the man's face that even though our training datasets do not contain any ground truth for people, we are able to successfully generate an accurate shading map with a sparse albedo.

\begin{figure*}
    \centering
    \includegraphics[width=\linewidth]{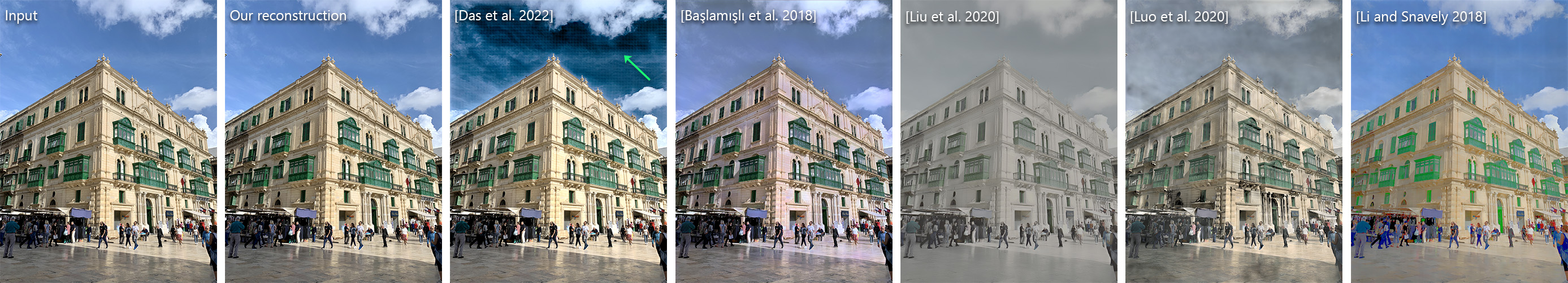}
    \caption{
    The reconstruction of the original image through the intrinsic equation in Equation ~\eqref{eq:intrinsicmodel} using the estimated albedo and shading maps is critical for an intrinsic decomposition method for down-the-line image editing tasks. 
    We show the reconstruction results of each method using the decompositions shown in Figure~\ref{fig:itw_main_comp}. 
    All the methods here other than ours use the intrinsic equation as a loss function and fail to faithfully reconstruct the original image. 
    As we directly infer our albedo using our estimated shading and the original image using the intrinsic equation, our albedo and shading estimations perfectly reconstruct the input image by definition. 
    \hfill \footnotesize{Image from Unsplash by William Jones.}
    }
    \label{fig:itw_main_comp_rec}
\end{figure*}

\paragraph{Reconstructing the original image}
Most intrinsic decomposition methods formulate separate albedo and shading estimation modules in their pipelines, using the intrinsic model in Equation~\ref{eq:intrinsicmodel} as a loss on the faithful reconstruction of the original image. 
As seen in Figure~\ref{fig:itw_main_comp_rec}, however, the final results do not necessarily satisfy this equality. 
USI and NIID-Net lose much of the original color content in their reconstruction, while CGIntrinsics loses contrast mainly due to their shading estimate, and JSI shifts the original colors in the image mostly due to their albedo estimate. 
While PIE-Net is able to reconstruct the image more faithfully than these methods, it may fail to reconstruct bright colors as seen in Figure~\ref{fig:introComparisonReconst} or create large artifacts as seen in the sky in Figure~\ref{fig:itw_main_comp_rec}. 
Our formulation that computes the estimated albedo directly from the shading does not suffer from reconstruction errors as the image is perfectly reconstructed by definition.
This is a critical property for intrinsic decomposition methods since this reconstruction informs any down-the-line image editing tasks.

We see similar performance of these methods in Figures~\ref{fig:itw_main_comp} and \ref{fig:itw_main_comp_rec} across a variety of photographs. 
We present these comparisons on 100 other images in our supplementary material. 

\definecolor{InRed}{rgb}{0.8,0,0}
\definecolor{InGreen}{rgb}{0,0.8,0}
\definecolor{InBlue}{rgb}{0.2,0.2,1}
\definecolor{InCyan}{rgb}{0,0.8,0.8}
\definecolor{InMagenta}{rgb}{0.8,0,0.8}
\definecolor{InYellow}{rgb}{0.8,0.8,0}

\newcommand{\inR}{\textcolor{InRed}{R}}
\newcommand{\inG}{\textcolor{InGreen}{G}}
\newcommand{\inB}{\textcolor{InBlue}{B}}
\newcommand{\inC}{\textcolor{InCyan}{C}}
\newcommand{\inM}{\textcolor{InMagenta}{M}}
\newcommand{\inY}{\textcolor{InYellow}{Y}}

In order to examine some of the typical challenges in intrinsic decomposition at a deeper level, we show 6 more examples in Figures~\ref{fig:itw_das_comp}-\ref{fig:itw_das_comp3}, comparing our method to that of Das et al.~\shortcite{das2022pie} (PIE-Net). 
The results of the other methods for these images are presented in the supplementary material. 
We named each of the examples in these figures and color-coded the insets (\texttt{\inR\inG\inB\inC\inM\inY}) for easy referral. 
For example, we will refer to the first inset in Figure~\ref{fig:itw_das_comp} as \texttt{(Attic)-\inR}.

\paragraph{Shadows on surfaces}
Since shadows are a direct result of illumination, we expect to see a smooth albedo and a sharp discontinuity in shading across shadow boundaries. 
The insets \texttt{(Attic)-\inR\inG} and \texttt{(Desk)-\inG\inM} show cast shadows on flat surfaces well-represented in our shading with a sparse corresponding albedo, while the albedos estimated by PIE-Net have the shadows still present. 
Similarly, we generate sparse albedo maps across the strong cast shadow boundaries on the leaves in \texttt{(Peppers)-\inB\inM} and on the wall and the ground in \texttt{(Tea)-\inR\inG\inY\inC\inM}. 
PIE-Net, on the other hand, generates low-contrast shading maps in such regions with shadows still visible in their estimated albedo.

\paragraph{Color shift under mixed illumination}
The single-channel shading model assumes a single-color illumination. 
In the presence of light sources with varying colors, such as light coming from outdoors and the ambient light present in the room in \texttt{(Desk)-\inB}, \texttt{(Tea)-\inR\inY}, and \texttt{(Kitchen)-\inR}, the color of the ambient light gets represented by a color shift in the albedo. 
Similarly, the shadows in indoor scenes are only illuminated with secondary illumination reflected from the surrounding surfaces. 
The color of this ambient illumination causes a color shift in the albedo as \texttt{(Attic)-\inY\inC\inM} and \texttt{(Chairs)-\inR} show. 
While this colored illumination is well-represented in our albedo estimations, we see that the albedos estimated by PIE-Net display a strong shift in albedo brightness \revision{while} representing the shadows incorrectly in their estimations.

\paragraph{Shading smoothness on primitive geometry}
We expect a smooth shading gradient on regions with simple geometry such as a plane or a cylinder. 
Despite the dramatic change in the albedo on the mug in \texttt{(Attic)-\inC} or on the ground in \texttt{(Kitchen)-\inM}, we are able to generate a smooth gradient across these surfaces. 
The geometry of the cup holder in \texttt{(Attic)-\inM} is also clearly visible in our shading estimation. 
On the contrary, in the shading estimations by PIE-Net, we see a strong shading discontinuity on the smooth ground in \texttt{(Kitchen)-\inM}, the pattern from the albedo still visible in \texttt{(Attic)-\inC}, and the shadows not well represented in \texttt{(Attic)-\inM}.

\paragraph{Albedo sparsity}
Single-colored objects should appear with a sparse albedo map free of shading variations in a successful intrinsic decomposition. 
As the \texttt{(Peppers)-\inR\inG\inB} under strong sunlight and \texttt{(Chairs)-\inR\inG\inB} under a soft ambient light show, we are able to generate sparse albedos for such objects under varying conditions. 
In these examples as well as in \texttt{(Desk)-\inG\inM} and \texttt{(Kitchen)-\inR}, the albedo maps estimated by PIE-Net incorrectly include the changes in shading, resulting in non-sparse albedos for flat-colored regions.

\paragraph{Specular surfaces}
Specular surfaces are challenging for intrinsic decomposition methods due to their very bright shading when compared to surfaces that can be modeled as Lambertian. 
Our inverse shading representation allows us to represent the wide range of shading values effectively by spreading the distribution of shading values in $[0,1]$. 
Our system can generate smooth shading maps of specular objects such as the lamp in \texttt{(Desk)-\inR}, the kitchen hood in \texttt{(Kitchen)-\inB}, or the metallic surfaces in \texttt{(Chairs)-\inB\inM}. 
We are also able to generate high-contrast shading maps for materials like leather as \texttt{(Tea)-\inB} and Figure~\ref{fig:iiw_dl_comp} show. 
We can also represent smaller specular regions such as on the glasses in \texttt{(Desk)-\inM} and on the legs of the chair in \texttt{(Chairs)-\inR}.
As also shown in Figures~\ref{fig:teaser} and \ref{fig:introComparison}, our method significantly improves on the state-of-the-art in representing specular surfaces.

\section{Quantitative Evaluation}
\label{sec:soa_comparison}

While qualitative comparison is the most reliable source of evaluation for intrinsic decomposition, we provide exhaustive quantitative evaluation on multiple existing benchmark datasets. We first present an evaluation on diverse synthetic scenes using the As Real as Possible (ARAP) Dataset~\cite{bonneel2017intrinsic}. We then perform real-world evaluations using the Intrinsic Images in the Wild (IIW) Dataset~\cite{bell2014intrinsic}, and the Shading Annotations in the Wild Dataset~\cite{kovacs17shading}. We also discuss multiple factors to consider when comparing these results. Finally, we give an analysis of the run-time and memory consumption of each method.

\subsection{ARAP Dataset}
\label{sec:arap_dataset}
In an attempt to quantify the physical accuracy and reconstruction of each approach, we perform quantitative comparisons on the synthetic ARAP Dataset \cite{bonneel2017intrinsic} in the zero-shot setting wherein our model was not trained on any data in the ARAP Dataset. We compare our approach to multiple recent deep learning approaches \cite{das2022pie, luo2020niid, liu2020unsupervised, li2018cgintrinsics, baslamisli2018joint}, two optimization-based approaches \cite{bell2014intrinsic, shen2011intrinsic}, and two naive baselines. We focus our analysis on these methods as they provide open-source code for their methods and predict single-channel shading like our approach.

\subsubsection{Evaluation Dataset}
The ARAP Dataset consists of approximately 150 rendered scenes varying in size, realism, and subject matter. Although the scenes are rendered, they allow for concrete quantitative comparisons since the ground-truth intrinsics are provided. It is important to note that some renders from this dataset are provided as part of the CGIntrinsics Dataset~\cite{li2018cgintrinsics}, therefore methods that train on CGIntrinsics cannot be considered zero-shot. This includes the works of~\citet{luo2020niid} and \cite{li2018cgintrinsics}. We still include these methods in our evaluation.

\begin{table}[]
\caption{Zero-shot comparison to prior work on the ARAP Dataset \cite{bonneel2017intrinsic}. We evaluate two naive baselines, two optimization-based methods, and five deep learning methods. Our proposed approach achieves the best performance across all metrics when considering zero-shot methods. We include two non-zero-shot methods that were trained on scenes in the ARAP dataset as part of the CGIntrinsics Dataset, namely \citet{li2018cgintrinsics} and \citet{luo2020niid}. We are able to perform competitively to these non-zero-shot methods, even outperforming them in shading prediction. Finally, by definition, our method has zero reconstruction error contrary to the other deep learning methods evaluated.}
\begin{adjustbox}{width=\linewidth}
{\renewcommand{\arraystretch}{1.2}}
\begin{tabular}{l|ccc|ccc|c}
\multirow{2}{*}{Method}                     & \multicolumn{3}{c|}{Shading}                     & \multicolumn{3}{c|}{Albedo}                         & \multicolumn{1}{c}{Recon.} \\
                                            & LMSE$\da$       & RMSE$\da$       & SSIM$\ua$    & LMSE$\da$       & RMSE$\da$       & SSIM$\ua$       & MSE$\da$                           \\ \hline
Chromaticity                                & 0.093 & 0.347 & 0.760 & 0.024 & 0.296 & 0.692 & --                              \\
Constant Shading                            & 0.124 & 0.403 & 0.634 & 0.046 & 0.454 & 0.663 & --                              \\ \hline
\cite{luo2020niid}*                         & 0.101 & 0.324 & 0.728 & 0.022 & 0.206 & 0.781 & 0.017                        \\
\cite{li2018cgintrinsics}*                  & 0.112 & 0.374 & 0.688 & 0.018 & 0.227 & 0.749 & 0.029                        \\ \hline
\cite{bell2014intrinsic}                    & 0.104 & 0.357 & 0.727 & 0.028 & 0.317 & 0.718 & \textbf{0.000}                        \\
\cite{shen2011intrinsic}                    & 0.093 & 0.355 & 0.742 & \textbf{0.021} & 0.313 & 0.705 & \textbf{0.000}                        \\
\cite{liu2020unsupervised}                  & 0.102 & 0.355 & 0.719 & 0.036 & 0.286 & 0.713 & 0.022                        \\
\cite{baslamisli2018joint}                  & 0.107 & 0.373 & 0.712 & 0.033 & 0.367 & 0.703 & 0.004                        \\
\cite{das2022pie}                           & 0.094 & 0.336 & 0.768 & 0.026 & 0.273 & 0.754 & 0.003                        \\ \hline
Ours (CGI only)                             & 0.090 & 0.337 & 0.757 & 0.025 & 0.262 & 0.751 & \textbf{0.000} \\
Ours                                        & \textbf{0.086} & \textbf{0.334} & \textbf{0.776} & \textbf{0.021} & \textbf{0.252} & \textbf{0.761} & \textbf{0.000}

\end{tabular}
\end{adjustbox}
\label{tab:arap}
\end{table}

\subsubsection{Evaluation Metrics}
Evaluating intrinsic decomposition approaches on dense ground-truth data can be difficult, oftentimes metrics do not reflect qualitative observations. Any metrics utilized must be scale-invariant given the inherent scale ambiguity of the intrinsic image formulation. \citet{grosse2009ground} observe that a regular scale-invariant mean squared error (si-MSE) is unforgiving since the incorrect attribution of a single edge to either shading or albedo can result in errors across large regions of the image. They instead propose an alternative metric, local mean squared error (LMSE), that computes si-MSE in overlapping windows. Following previous works~\cite{fan2018revisiting, cheng2018intrinsic, das2022pie}, we utilize LMSE, scale-invariant root mean squared error (RMSE), and structural-similarity (SSIM). In addition to measuring the accuracy of each intrinsic component, we also measure the scale-invariant reconstruction error of each approach. 

\subsubsection{Quantitative Analysis} 
The quantitative results of our evaluation are shown in Table \ref{tab:arap}. Our method achieves the best performance across the board when considering zero-shot methods. Although the works of \citet{li2018cgintrinsics} and \citet{luo2020niid} train on many of the images from the ARAP Dataset, we still perform competitively with these methods. We are even able to outperform them in all metrics on shading prediction. Due to our formulation, our approach also has zero reconstruction error by definition. All other data-driven methods result in some reconstruction error. 

Table \ref{tab:arap} also shows the results of our pipeline when trained solely on the CGIntrinsics Dataset \cite{li2018cgintrinsics}, without the extra examples from the ARAP dataset. The CGIntrinsics Dataset consists of only approximately 20,000 examples of rendered indoor scenes. Despite this, our model is still able to achieve state-of-the-art results. Just like our full pipeline, we are even able to outperform the non-zero-shot methods in shading estimation. This exhibits the fact that a main contributing factor of our proposed approach is the careful design of our two-step shading estimation pipeline.

\begin{table}[]
\caption{Quantitative results on the IIW Dataset \cite{bell2014intrinsic} and the SAW Dataset \cite{kovacs17shading}. Our model is not trained, or fine-tuned on the IIW Dataset and therefore performs poorly contrary to qualitative observations. We show that by adding a constant 0.5 to our results we can achieve a state-of-the-art zero-shot score. For the SAW Dataset, our model achieves competitive results without training on it.
}
\resizebox{0.91\columnwidth}{!}{
\begin{tabular}{l c c c}
{Method} & {Training Set} & {WHDR} & {AP} \\ \hline
\cite{grosse2009ground} & - &  26.9 & 85.2 \\ 
\cite{garces2012intrinsic} & - &  24.8 & 92.3 \\ 
\cite{zhao2012closed} & - & 23.8 & 92.1 \\ 
\cite{bell2014intrinsic} & - & 20.6 & 89.7 \\ 
\hline
\cite{zhou2015learning} & IIW & 19.9 & 86.3 \\ 
\cite{liu2020unsupervised} & CGI + IIW & 18.6 & 85.3 \\
\cite{zhou2019glosh} & SUNCG + IIW + SAW & 15.2 & 95.0 \\
\cite{li2018cgintrinsics} & CGI + IIW + SAW & 15.5 & 97.9 \\
\hline
\cite{li2018cgintrinsics} & CGI & 17.8 & 94.0 \\ 
\cite{zhou2019glosh} & SUNCG & 26.8 & 92.4 \\
\cite{baslamisli2018joint} & NED & 38.0 & 82.4 \\
\cite{luo2020niid} & CGI + NYU & 16.6 & 98.4 \\ 
\cite{das2022pie} & NED & 21.3 & 82.8 \\ \hline
Ours & CGI + GTA + OR + HS + MI & 24.9 &  95.5 \\ 
Ours + 0.5 & CGI + GTA + OR + HS + MI & 15.3 & --
\end{tabular}
}
\label{tab:iiw}
\end{table}

\subsection{IIW Dataset}
Many prior works use the pair-wise reflectance judgments of the IIW dataset to evaluate the quality of their inferred albedo components. The dataset was introduced with a metric called weighted human disagreement rate (WHDR). This metric measures the rate at which predicted ordinal relationships match the ground truth annotations, weighted by a confidence computed based on the consistency of human annotations. It has been shown that the WHDR metric of the IIW Dataset does not reflect the true physical accuracy or editing capability of intrinsic decomposition approaches \cite{garces2022survey}. An accurate comparison of methods on the IIW dataset is complicated given that various methods perform training and/or validation on the IIW Dataset. For thoroughness, we provide our quantitative results for the IIW Dataset without any training occurring on images from the IIW Dataset. We follow the commonly used testing split proposed by \citet{narihira2015learning}. 

Table \ref{tab:iiw} shows the results of various prior works on the IIW test split. Although WHDR scores have trended downward with the introduction of data-driven methods, reported scores vary drastically depending on whether or not methods are trained on the IIW training split. Naturally, methods that finetune on the IIW Dataset achieve significantly lower scores than those that do not. While some methods may not directly train on the IIW Dataset, they may use the training split as validation when early-stopping their training, or to drive the design choices of their methods, resulting in noticeably lower scores.

\begin{figure*}[]
    \centering
    \includegraphics[width=\linewidth]{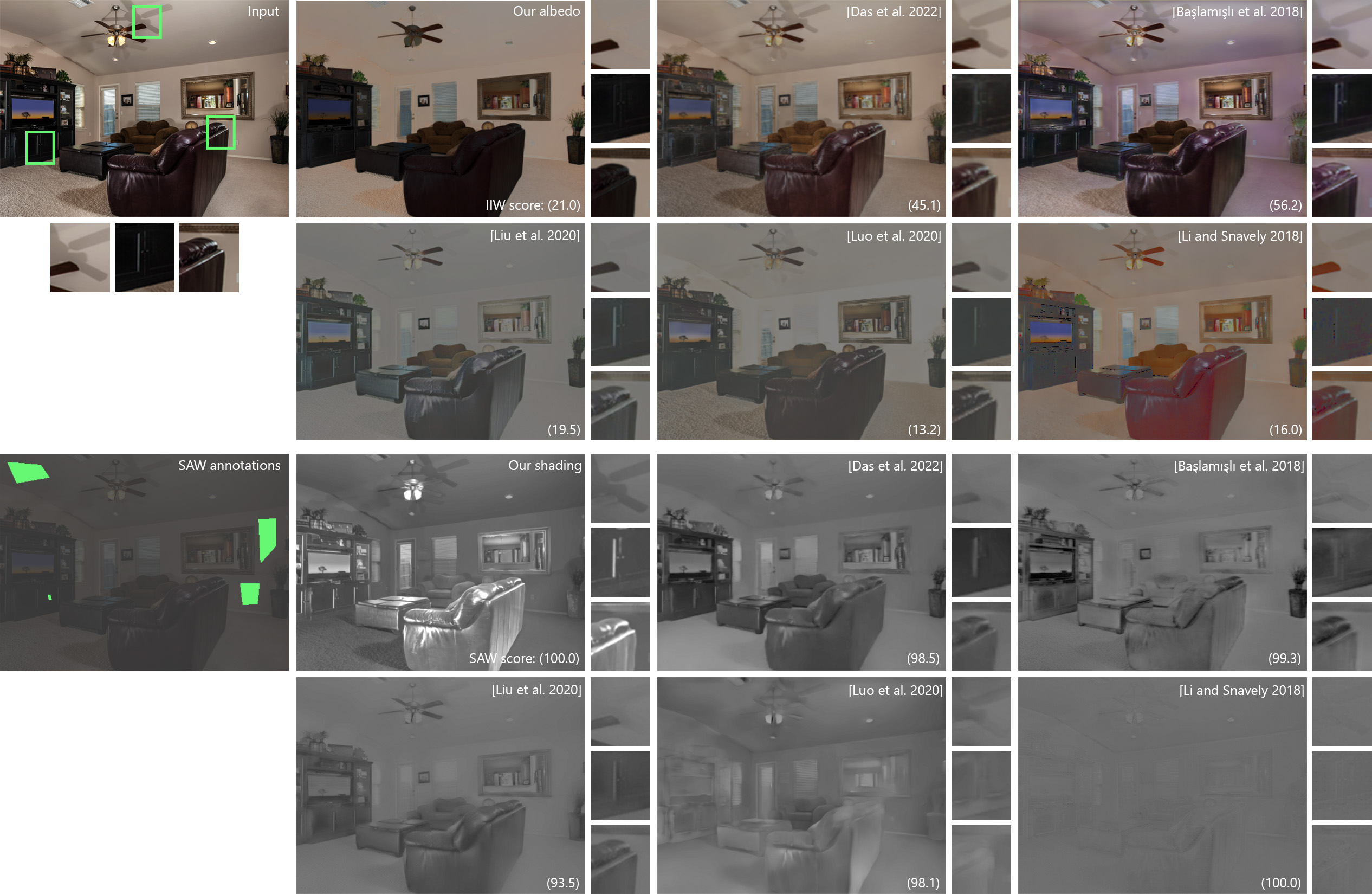}
    \caption{
    Qualitative comparison of quantitative metrics. 
    We present two error metrics commonly utilized to measure intrinsic decomposition performance on this example. 
    (IIW) Together with the albedo estimations by state-of-the-art methods, we report the weighted human disagreement rate (WHDR) sparse error metric defined on the IIW dataset~\cite{bell2014intrinsic} where lower numbers are better. 
    Despite the washed-out colors and albedo sparsity issues in challenging regions as shown in the insets, the three methods on the bottom row score favorably compared to ours. 
    (SAW) We show the regions in which the shading smoothness is evaluated on the left and report the AP scores defined on the SAW dataset~\cite{kovacs17shading} together with the shading estimations, where higher numbers are better. 
    We see that most competing methods generate very smooth shading estimations, which results in very high scores on this metric despite the lack of contrast and inconsistent shading across different surfaces as shown in the insets. 
    Our method can generate high-contrast shading estimations while achieving very high scores on this metric. 
    This shows a mismatch between these two error metrics and the qualitative performance of intrinsic decomposition algorithms, as also discussed by Garces et al.~\shortcite{garces2022survey}. 
    \hfill \footnotesize{Image from Flickr by Bill Wilson.}
    }
    \label{fig:iiw_dl_comp}
\end{figure*}

Rather than this change in score reflecting generalization to real-world scenes, we believe there may be easily exploitable aspects of the WHDR metric. For example, \citet{nestmeyer2017reflectance} show that by simply scaling the input images into the range $[0.55, 1]$, a score of $25.7$ can be achieved due to the imbalance of equal and non-equal annotations. Similarly, we show that by simply shifting our \revision{albedo estimations} by $0.5$ we are able to achieve state-of-the-art performance among zero-shot approaches. \revision{Note the shifted albedo values are not constrained to be in the $[0,1]$ range.} Methods trained on the IIW Dataset may use this imbalance to produce lower scores making it difficult to compare approaches. This is further exemplified in quantitative results shown in Figure \ref{fig:iiw_dl_comp}. It is unclear what causes certain methods to yield lower scores, despite perceptually worse albedo estimations. 

\subsection{SAW Dataset}
We additionally evaluate our proposed approach on the SAW Dataset \cite{kovacs17shading}. The SAW Dataset consists of annotations of shading smoothness and discontinuities on images from the IIW and the NYU Depth Dataset V2. Given a shading estimation, each region can be classified as smooth or non-smooth based on image gradients. The authors compute an average precision metric by computing the classification precision of these regions over varying gradient threshold values. We utilize the improved challenge metric proposed by \cite{li2018cgintrinsics} that weighs smooth shading regions proportionally to how difficult they are to predict properly. Since many smooth shading regions occur over smooth regions of the image, this improved metric computes the average image gradient over each smooth shading region to determine its difficulty. 

Similar to the WHDR metric, the SAW AP metric only gives a partial evaluation of decomposition quality. Since the AP metric only measures smooth regions and discontinuities, methods can achieve high scores with low contrast shading estimations as long as a measurable amount of gradient is predicted at discontinuities. The metric does not measure whether or not shading discontinuities have the correct magnitude. Nevertheless, our method performs very competitively, outperforming all other zero-shot methods besides \citet{luo2020niid} which uses the SAW training set as validation to evaluate \revision{the performance of their model} while training.

\begin{figure*}
    \centering
    \includegraphics[width=\linewidth]{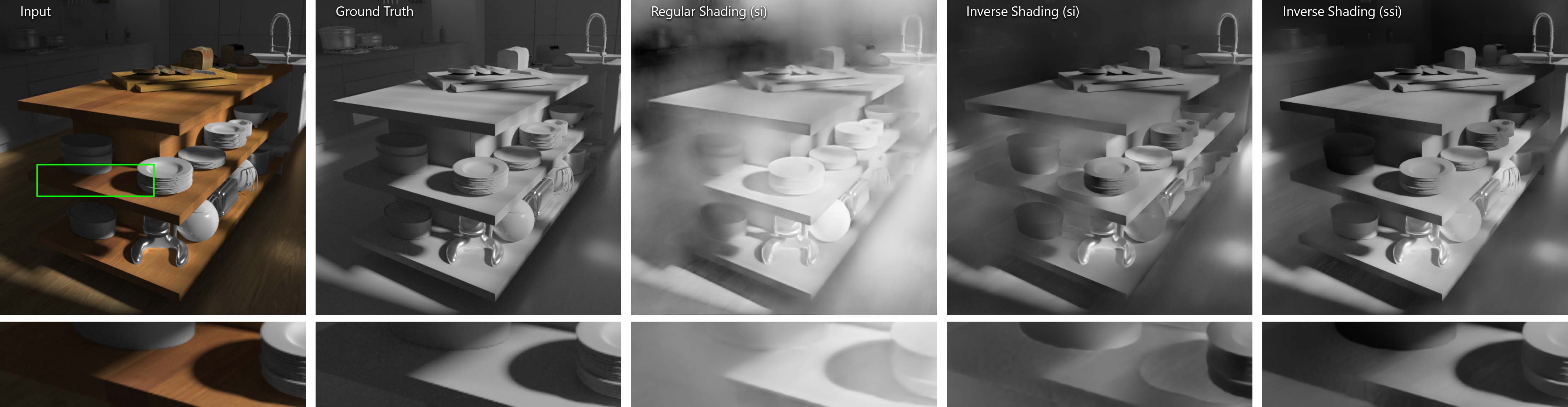}
    \caption{An example from our controlled experiment comparing ordinal training strategies on the ARAP Dataset. Training the network in the regular shading space with a scale-invariant loss yields ordinal estimations that lack contrast and suffer from global inconsistencies. By training in our proposed inverse shading space, the network generates estimations with higher contrast and improved global structure. Finally, by further relaxing the training formulation with a shift- and scale-invariant loss, we are able to generate highly-detailed, accurate ordinal constraints.}
    \label{fig:exp:ordinalComparison}
\end{figure*}

\subsection{Run-Time Analysis}

In order to effectively utilize intrinsic decomposition for image editing applications, a given algorithm needs to be efficient in both time and memory. Taking multiple minutes of run-time or requiring a high-end GPU can greatly limit the usability of an intrinsic decomposition approach. 

Our pipeline consists of two networks, our ordinal network being larger than our intrinsic decomposition network. Given our local and global ordinal estimation formulation, the image must be run through the first network twice, once at the training resolution and a second time at the $\mathcal{R}_0$ resolution. These two estimations are then fed to the second network, along with the input image to compute our final shading estimation.  

\revision{All timing and memory consumption measurements are performed using a machine with an RTX 2060 Super GPU, an Intel i5-9600k CPU, and 16 gigabytes of RAM.}
It takes our method $0.3$ seconds in total to process a $(768 x 768)$ image. The works \citet{liu2020unsupervised}, \citet{luo2020niid}, and \citet{li2018cgintrinsics} take $0.06$, $0.08$ and $0.005$ seconds respectively. This discrepancy in runtime is likely due to the multiple forward passes that are required in our pipeline. The work of \citet{das2022pie} takes $8$ seconds for the same image. 

As for memory efficiency, our method uses $1.2$ gigabytes of GPU memory for an image of size $(768 x 768)$. The most memory-efficient network is that of \citet{li2018cgintrinsics} which requires about $0.7$ gigabytes for the same size image. The networks of \citet{liu2020unsupervised}, \citet{luo2020niid} and \citet{das2022pie} all require more memory than our method, using $1.3$, $2.5$ and $8$ gigabytes, respectively.

\section{Ablation Studies}
\label{sec:ablations}
To evaluate the various design choices of our proposed pipeline, we carry out multiple controlled experiments and perform zero-shot evaluations using the ARAP Dataset described in Section \ref{sec:arap_dataset}. In addition to the ablations of the main paper, we provide additional discussion of our multi-illumination training strategy in the supplementary material.

\begin{figure*}[]
    \centering
    \includegraphics[width=\linewidth]{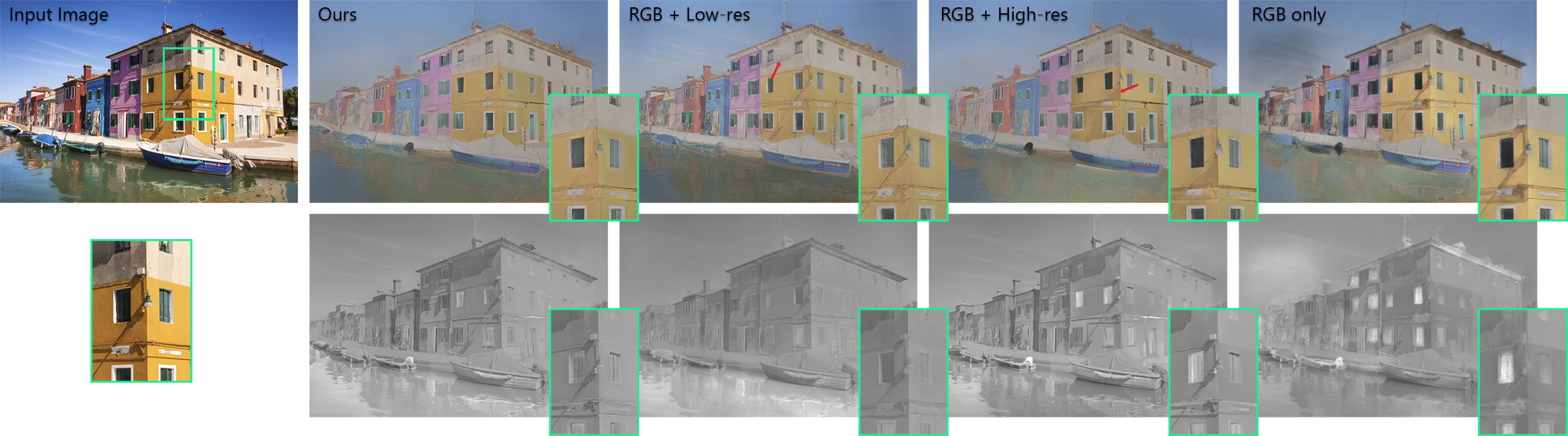}
    \caption{\revision{We perform an ablation over inputs to the second network. When only provided with the RGB image, the network has to perform the entire task of intrinsic decomposition, causing very noticeable artifacts and inconsistencies (far right column), showing the efficacy of our two-step approach. When provided with only high- or low-resolution ordinal inputs the network either misses sharp details (middle left column) or fails to predict globally coherent structure (middle right column). Our proposed multi-resolution input configuration generates the most accurate and coherent shading and albedo estimations. 
    \text{ } \hfill \footnotesize{Image from Unsplash by Dirk Sebregts.}
    }}
    \label{fig:res_abl}
\end{figure*}

\begin{figure*}
    \centering
    \includegraphics[width=\linewidth]{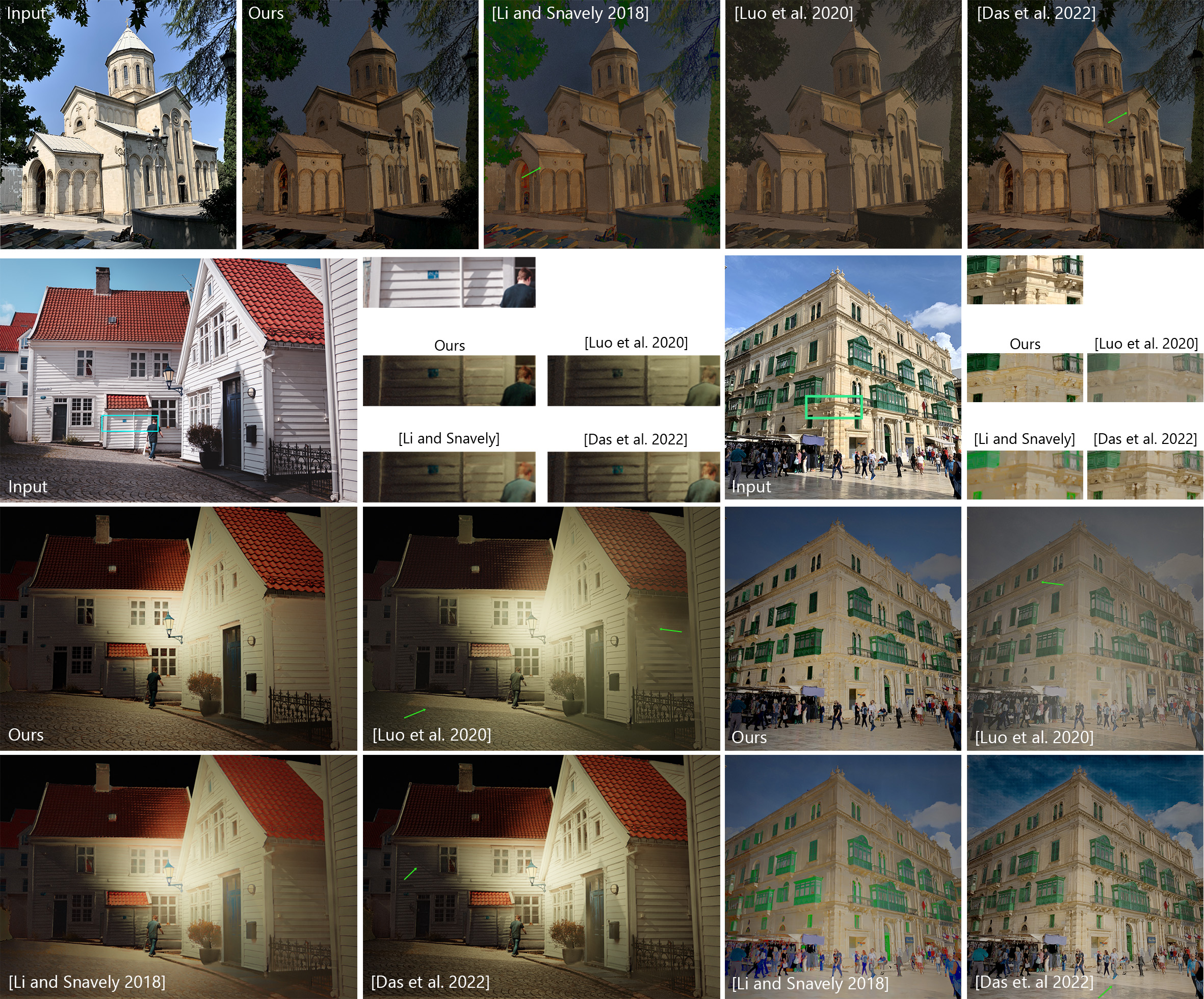}
    \caption{We utilize predicted albedo and surface normals to relight various scenes using competing intrinsic decomposition approaches. Our method is able to completely remove lighting effects from the original image. In the top row, we are able to remove the hard shadows on the church while other methods leave residual shadows, or shift the colors of the scene. The same behavior can be seen on the houses in the bottom left. Our method removes the shadows from the ground and the side of the houses, resulting in a more realistic relighting effect. In the building example, our albedo allows for accurate relighting of the building from the viewpoint of the camera. Other methods leave shadows on the left facade of the building. 
    \\ \text{  }
    \hfill \footnotesize{All images from Unsplash, bottom left by Lieuwe Terpstra and bottom right by William Jones.}
    }
    \label{fig:app_relight}
\end{figure*}

\begin{figure*}
    \centering
    \includegraphics[width=\linewidth]{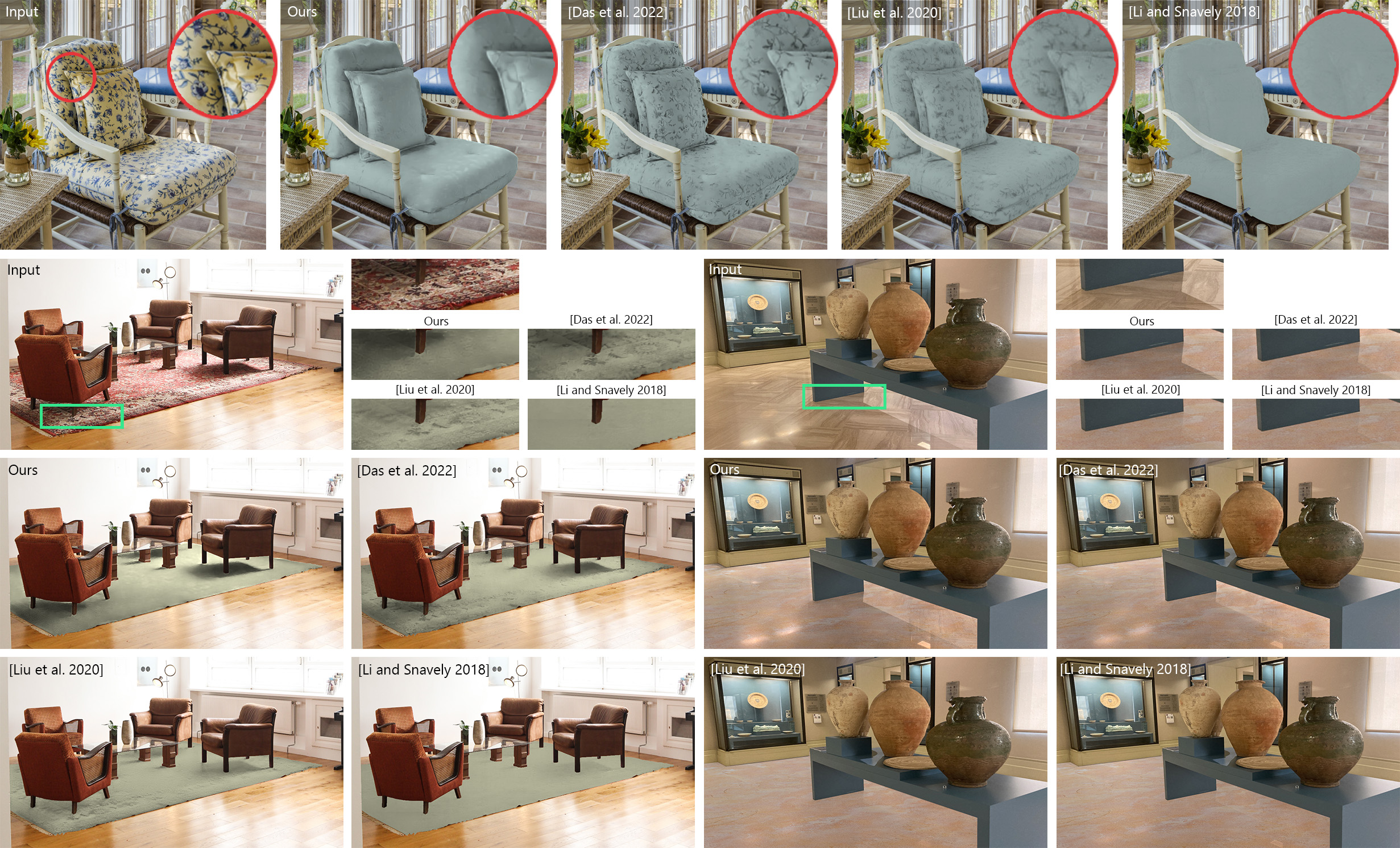}
    \caption{We make use of our decompositions to create realistic illumination-aware recoloring edits. In the top row, we are able to recolor the underlying albedo of the yellow chair to a solid color. Other methods can not fully separate albedo and shading resulting in albedo textures leaking into the final edit. Similarly, we can recolor the patterned rug in the bottom right, while other methods leak albedo texture, or fail to accurately represent shading effects. In the bottom right example, our method can disentangle specular lighting effects from albedo allowing for realistic recoloring of shiny surfaces. 
    \\ \text{  }
    \hfill \footnotesize{Two images from Unsplash by Francesca Tosolini (top) and Beazy (bottom left).}
    }
    \label{fig:app_recolor}
\end{figure*}

\subsection{Analysis of Ordinal Training}
\label{sec:ablation:ordinal}
We evaluate the efficacy of our ordinal training formulation using a controlled experiment. We compare three strategies for learning ordinal shading estimation; regular shading with a scale-invariant loss, inverse shading with a scale-invariant loss and inverse shading with a scale- and shift-invariant loss.

\begin{table}[]
\caption{We find that our inverse shading representation with a scale-invariant loss generates more globally consistent ordinal estimations when compared to using the same loss in the shading space. Furthermore, our inverse representation paired with a shift- and scale-invariant loss yields the best results both globally and locally, indicated by the ordinal and D$^3$R metrics, respectively.}
\begin{adjustbox}{width=0.7\linewidth}
\begin{tabular}{l|ll|ll}
\multicolumn{1}{c|}{\multirow{2}{*}{Method}} & \multicolumn{2}{c|}{384} & \multicolumn{2}{c}{$\mathcal{R}_0$} \\
\multicolumn{1}{c|}{}                        & Ord$\downarrow$  & D$^3$R$\downarrow$     & Ord$\downarrow$ & D$^3$R$\downarrow$ \\ \hline
SI-shading                                   & 0.377             & 0.246               & 0.386            & 0.260           \\
SI-inverse                                   & 0.318             & 0.230               & 0.320            & 0.307           \\
SSI-inverse                                  & \textbf{0.294}    & \textbf{0.207}      & \textbf{0.297}   & \textbf{0.231} %
\end{tabular}
\end{adjustbox}
\label{tab:ord_train}
\end{table}

\subsubsection{Training Setup}
We utilize subsets of the Hypersim \cite{roberts2021hypersim} and GTA \cite{krahenbuhl2018free} datasets for training. We sample 5000 images from each dataset's training split and train each approach for 100 epochs. We include the gradient loss described in Equation \ref{eq:method:ord_grad_loss} in each experiment.

\subsubsection{Evaluation Setup}
To evaluate the generalization capabilities of each approach, we perform zero-shot evaluation on the ARAP dataset \cite{bonneel2017intrinsic}. We use existing ordinal metrics borrowed from the monocular depth estimation \revision{literature} to compare the effectiveness of each training formulation. The pair-wise ordinal metric (Ord.)~\cite{xian2020structure} measures the correctness of ordering between randomly sampled pixels. The D3R metric~\cite{miangoleh2021boosting} similarly measures ordinality across discontinuities determined by the super-pixel segmentation of the ground-truth shading. These roughly quantify global ordinality and local ordinality, respectively. We evaluate at two resolutions; 384 px and $\mathcal{R}_0$

\subsubsection{Analysis}
Table \ref{tab:ord_train} shows the resulting ordinal metrics across the various training setups. The first row shows the typical intrinsic decomposition training strategy wherein the network directly regresses values in the original shading space using a scale-invariant loss (si-MSE). The second row shows that by learning the shading in our transformed inverse space, the network is better at generating globally coherent ordinal estimations, but the D3R metric indicates worse local performance at high-resolution. Finally, our proposed formulation instead uses a scale- and shift-invariant loss (ssi-MSE) in the inverse shading space. This approach performs the best in both metrics at both resolutions indicating that it is capable of generating accurate global structure as well as local discontinuities. We also attempted to train a fourth network using ssi-MSE in the original shading space but this network diverged. This is likely due to a poor distribution of values, as demonstrated in Figure~\ref{fig:ordinal:shd_rep}, combined with the shift- and scale-invariant loss function. 

\begin{table}[]
\caption{Results of our controlled experiment comparing the efficacy of our loss formulation when training our second network. Adding losses to the implied albedo significantly improves the performance on albedo estimation and even improves the performance on shading estimation, indicating that each intrinsic component provides complementary supervision. The ARAP Dataset is used for evaluation at 1024-pixel resolution}
\begin{adjustbox}{width=\linewidth}
{\renewcommand{\arraystretch}{1.5}}

\begin{tabular}{lllllll}
\multicolumn{1}{c|}{\multirow{2}{*}{Method}} & \multicolumn{3}{c|}{Shading}            & \multicolumn{3}{c}{Albedo} \\
\multicolumn{1}{c|}{}                        & LMSE               & RMSE               & \multicolumn{1}{l|}{SSIM}               & LMSE             & RMSE            & SSIM        \\ \hline
\multicolumn{1}{l|}{Shd.}                    & 0.0680             & 0.3275             & \multicolumn{1}{l|}{0.7854}             & 0.0191           & 0.2783          & 0.7645       \\
\multicolumn{1}{l|}{Shd. + Alb.}             & \textbf{0.0675} & \textbf{0.3217} & \multicolumn{1}{l|}{\textbf{0.7889}} & \textbf{0.0173}  & \textbf{0.2661} & \textbf{0.7780}      
\end{tabular}

\end{adjustbox}
\label{tab:met_abl}
\end{table}

\subsection{Analysis of Joint Albedo and Shading Loss}
\label{sec:ablation:jointloss}
We show the effectiveness of utilizing losses on both estimated shading and the corresponding implied albedo by performing a controlled ablation study with and without albedo loss. 

\subsubsection{Training Setup}
We train using a subset of the Hypersim \cite{roberts2021hypersim} dataset consisting of 10,000 examples. We use our fully-trained ordinal network to generate input estimations and only train our second network. The two variants are both trained for 100 epochs with a learning rate of $10^{-5}$

\begin{table}[]
\caption{\revision{Results of our controlled experiment comparing input configurations for the second network. Providing the network with both low-resolution and high-resolution ordinal input yields the best performance, especially on global scale-invariant networks. Providing ordinal input significantly improves performance when compared to only feeding the network the RGB image as input, highlighting the efficacy of our two-step approach.}}
\begin{adjustbox}{width=\linewidth}
{\renewcommand{\arraystretch}{1.5}}
\begin{tabular}{lllllll}
\multicolumn{1}{c|}{\multirow{2}{*}{Inputs}} & \multicolumn{3}{c|}{Shading}            & \multicolumn{3}{c}{Albedo} \\
\multicolumn{1}{c|}{}                        & LMSE$\da$               & RMSE$\da$  & \multicolumn{1}{l|}{SSIM$\ua$}               & LMSE$\da$             & RMSE$\da$            & SSIM$\ua$        \\ \hline
\multicolumn{1}{l|}{All}                    & \textbf{0.0620}    & \textbf{0.3100} & \multicolumn{1}{l|}{\textbf{0.7808}} & 0.0180  & \textbf{0.2665} & \textbf{0.7650}       \\
\multicolumn{1}{l|}{RGB + Full}             & \textbf{0.0620}    & 0.3323 & \multicolumn{1}{l|}{0.7660} & 0.0187  & 0.2866 & 0.7575  \\
\multicolumn{1}{l|}{RGB + Base}             & 0.0633             & 0.3157 & \multicolumn{1}{l|}{0.7651} & \textbf{0.0178}  & 0.2766 & 0.7631  \\
\multicolumn{1}{l|}{RGB}                    & 0.0645             & 0.3554 & \multicolumn{1}{l|}{0.7467} & 0.0197  & 0.3368 & 0.7476  \\
\end{tabular}

\end{adjustbox}
\label{tab:res_abl}
\end{table}

\subsubsection{Evaluation Setup}
We perform zero-shot evaluation on the previously described ARAP Dataset. We evaluate the final estimations using three scale-invariant metrics. Specifically, we evaluate scale-invariant root mean squared error (RMSE), LMSE, and SSIM on both the predicted shading and albedo components. Given our focus on high-definition estimations, we evaluate each training formulation at the 1024-pixel resolution. 

\subsubsection{Analysis} Table \ref{tab:met_abl} shows the results of the joint loss ablation experiment. The first row shows the result of only computing losses on the shading component during training. As expected, adding losses on the implied albedo significantly improves \revision{the performance of the model} on albedo estimation. Furthermore, we observe that the albedo loss even slightly improves the accuracy of our shading estimation, suggesting that these losses provide complementary forms of supervision. Additionally, we find that when the network is only trained using the losses on the shading component, our pipeline produces artifacts in image regions with low shading values. Some examples of this behavior are shown in Figure \ref{fig:highdef_loss_comp}. The artifacts are alleviated by the addition of the albedo loss, further exemplifying the harmonious nature of these two supervision sources.

\subsection{Multi-Resolution Ablation}
We show the effectiveness of our proposed multi-resolution ordinal estimations by performing a controlled experiment over various possible input configurations of our second network.

\subsubsection{Training Setup}
We train each variant using the Hypersim \cite{roberts2021hypersim} dataset. We use our fully-trained ordinal network to generate input estimations and only train our second network. The four variants are trained for 100,000 iterations with a learning rate of $10^{-5}$ and a batch size of 8.

\subsubsection{Evaluation Setup}
We perform zero-shot evaluation on the ARAP Dataset. We evaluate the final estimations using three scale-invariant metrics. Specifically, we evaluate scale-invariant root mean squared error (RMSE), LMSE and SSIM on both the predicted shading and albedo components. Since our multi-resolution approach allows us to predict consistent estimations at high resolution, we perform our evaluation at our previously described $\mathcal{R}_0$ resolution.

\subsubsection{Analysis}
Table \ref{tab:res_abl} shows the results of the ablation. The first row shows the result of our proposed approach where we provide the second network with the RGB image, and ordinal estimations at both the base resolution and the full resolution. The two following rows show our method excluding one of these two ordinal estimations. We observe that by excluding either of the ordinal estimations, our performance considerably degrades, especially on the more difficult globally scale-invariant metrics RMSE, and SSIM. Furthermore, the final row shows the performance of our second network if we simply train it without any ordinal estimations, providing the network with only the RGB image. This configuration very significantly decreases our performance across all metrics. 

We also qualitatively observe noticeable differences between each training setup as shown in Figure \ref{fig:res_abl}. When only provided with the low-resolution ordinal estimation, the model isn't able to accurately predict sharp details in the shading layer, resulting in inaccurate predictions on small shadows (e.g. the thin shadow in the inset). When only provided with the high-resolution ordinal estimation, the model can generate detailed shading predictions but with global inconsistencies across distance image regions (on the water and on the building). When no ordinal estimation is used, the model consistently fails to predict accurate shading. Our full approach is able to generate very detailed estimations that are also globally coherent at high resolutions due to our multi-resolution approach.

\color{black}

\section{Applications}
\label{sec:applications}
Following the findings of~\citet{bonneel2017intrinsic} we evaluate the accuracy of our decompositions relative to state-of-the-art via intrinsic image editing comparisons. We perform two types of edits that are difficult to perform without access to intrinsic components, namely, illumination-aware recoloring, and single-image relighting.

\subsection{Relighting}
Since albedo is a representation of the scene without any lighting effects, it is an essential component for relighting. In order to show the usability of our decompositions, we perform relighting using our albedo and an off-the-shelf surface normal estimator~\cite{eftekhar2021omnidata}. \revision{We render the relighting results by loading our albedo and the estimated normals into Blender and use a standard diffuse shader to generate a novel shading layer. We insert virtual point lights into the scene to simulate various lighting conditions.}

Some examples of our relighting effect are shown in Figure~\ref{fig:app_relight} compared against \revision{other approaches}. \revision{Thanks} to our accurate albedo, our method is able to generate relit images without residual shading effects from the original illumination. In the top row, \citet{das2022pie} and \citet{li2018cgintrinsics} leave shadows on the building and the albedo of~\citet{luo2020niid} yields a relit image with low contrast. In the bottom left example, we can see that our method is able to remove the hard shadows on the sides of the houses and on the street. Finally, we show an example from Figure~\ref{fig:itw_main_comp}. The relit image is generated by placing a light source as if it was coming directly from the camera, therefore each face of the building should have the same brightness. Competing methods fail to remove the shadow from the left side of the building, resulting in inaccurate relit images. 

\begin{figure}[]
    \centering
    \includegraphics[width=\linewidth]{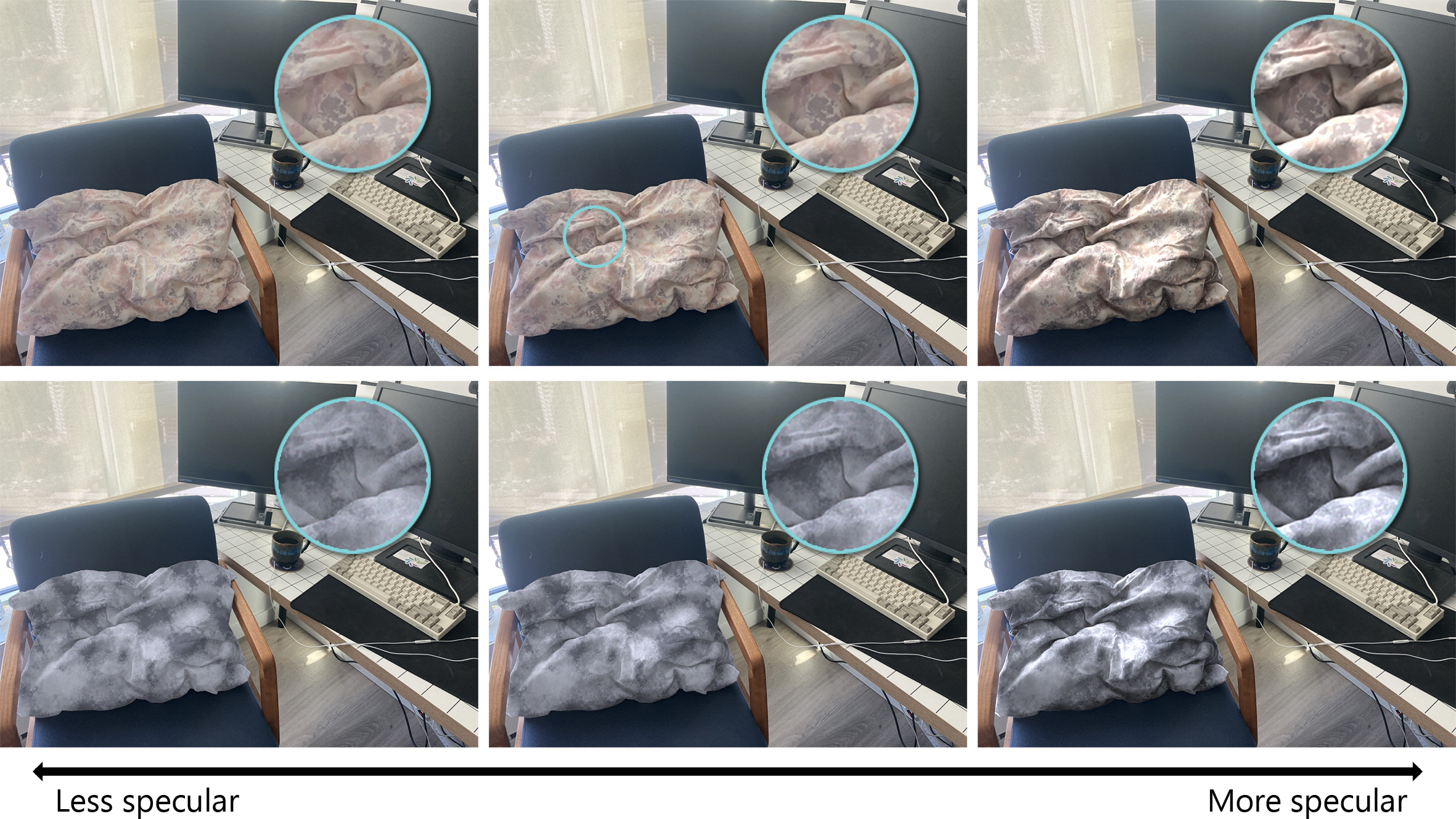}
    \caption{In addition to altering the albedo, we can alter the predicted shading values to simulate different material properties following \cite{ye2014intrinsic}. From left to right we show the same image with varying amounts of specularity. The bottom row shows the image with the edited albedo.}
    \label{fig:app_material}
\end{figure}

\subsection{Recoloring}
Simple recoloring edits can be made using intrinsic components. By altering the colors of certain regions in the albedo and reapplying the shading, we can edit surfaces while maintaining lighting effects. In the case of solid color objects, this kind of edit can be trivially achieved by decomposing the image into its chrominance and luminance \cite{bonneel2017intrinsic}. When both albedo and shading vary in a given region, the two components must be disentangled to perform recoloring. 

To generate the recoloring results, we compute estimated shading and albedo for each approach. Since most methods do not enforce image reconstruction, we use the shading and the input image to compute the implied albedo using Equation \ref{eq:intrinsicmodel}. We use this albedo for recoloring so that each method reconstructs the image and can be compared fairly. We show multiple examples of this type of edit in Figure \ref{fig:app_recolor} and compare them to state-of-the-art approaches. Our method is able to accurately separate lighting effects even when the albedo varies greatly in the edited region. The yellow chair and red carpet examples show that other methods either do not faithfully represent lighting effects or exhibit residual albedo patterns leaking into their shading component. In the museum example, our method, unlike others, is able to represent subtle specular effects that significantly increase the realism of the edit.

In addition to recoloring surfaces, we can also perform simple material editing by altering the distribution of estimated shading values. Figure~\ref{fig:app_material} shows an example of this type of edit with the original albedo and recolored albedo. By exponentiating the shading, the values can be compressed or expanded causing the edited surface to appear more, or less specular.

\section{Limitations}
\label{sec:limitations}

\begin{figure}[]
    \centering
    \includegraphics[width=\linewidth]{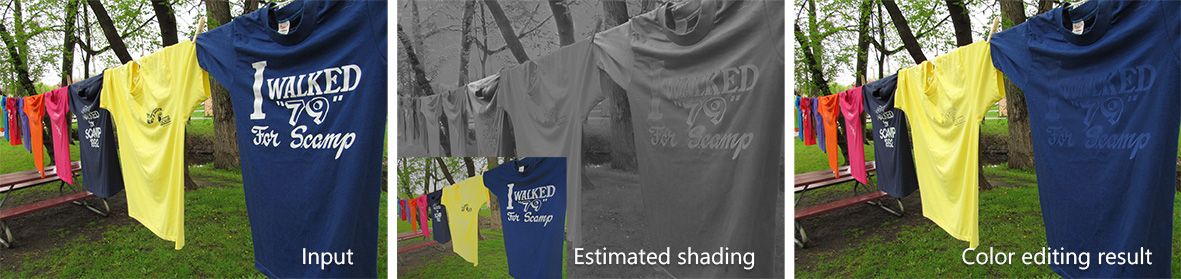}
    \caption{\revision{The Lambertian shading assumption adversely affects the shading smoothness across mixed materials. 
    \hfill \footnotesize{Image from Flickr by clarkstonscamp.}
    }}
    \label{fig:limitation}
\end{figure}

\revision{The main limitation of our method comes from the commonly-used gray-scale and Lambertian shading assumptions. 
While these assumptions make the complex intrinsic decomposition problem more tractable, it fails to represent various phenomenon commonly observed in real-world photographs. 
As discussed in Section~\ref{sec:itw_comp}, the gray-scale shading assumption results in shifted albedo colors in regions with multi-colored illumination, including the hard shadows where the illumination is dominated by secondary reflections. 
While our method is able to represent the specularities in the estimated shading, the presence of specularities adversely effects image editing results which relies on the Lambertian shading assumption. 
This assumption also affects the smoothness of the estimated shading on objects composed of different materials as seen on the t-shirt in Figure~\ref{fig:limitation}.}
Furthermore, since our method makes use of pseudo-ground truth, any biases present in the model's estimation may not be fully eliminated after computing our median albedo estimate. We typically observe mistakes in high-frequency albedo regions wherein the model may leave small gradients in the shading.

\section{Conclusion and Discussion}
\label{sec:conclusion}

\revision{
In this work, we present a new approach to intrinsic image decomposition in the wild. Through a comprehensive qualitative evaluation, we demonstrate that we can achieve high-quality and high-resolution intrinsic decomposition that allows the editing of illumination and material recoloring. 
Our main contribution is the introduction of the dense ordinal shading representation that simplifies the task and allows us to generate highly detailed shading discontinuities.
We estimate the ordinal shading in two resolutions and use them as input to the full intrinsic decomposition that makes it possible to regress the result at high resolutions.
We also propose to estimate the shading layer in the inverse shading domain that allows us to properly represent the specularities as well as the dark regions in the shading layer within $[0,1]$. 
We generate real-world training data using a robust albedo estimator from a multi-illumination dataset. By training on the generated dataset, we are able to bridge the domain gap between real-world and synthetic images. Additionally, despite the dataset consisting of indoor scenes, our method generalizes to diverse image content such as humans and outdoor scenes. 
}

\revision{
We evaluated our method qualitatively, as well as quantitatively using the commonly used benchmarks. 
As also widely discussed in the literature, our experiments show that the quantitative metrics and datasets used in the literature fail to reflect the performance of intrinsic decomposition methods. 
While the sparse annotations on real-world images provided in IIW and SAW have been useful for the community in the past, we believe that they are not reliable indicators to evaluate decomposition performance. 
Our pseudo-ground-truth generation method from multi-illumination images is a promising direction to formulate dense evaluation metrics to compare intrinsic decomposition methods. 
Currently, we believe that an extensive qualitative evaluation is necessary to demonstrate the advantages and disadvantages of new methodologies in this domain.
}

\section*{Acknowledgements}

We would like to thank our lab members Sebastian Dille for his feedback on the text, S. Mahdi H. Miangoleh for his support with the implementation, and Ari Blondal and Sepideh Sarajian Maralan for their help during the early stages of this work. We acknowledge the support of the Natural Sciences and Engineering Research Council of Canada (NSERC), [RGPIN-2020-05375]. 

\bibliographystyle{ACM-Reference-Format}
\bibliography{references}

\end{document}